\pdfoutput=1

\documentclass{template}

\usepackage{moreverb}
\usepackage[hyphens]{url}
\usepackage[colorlinks,bookmarksopen,bookmarksnumbered,citecolor=red,urlcolor=red]{hyperref}
\usepackage{siunitx}
\usepackage[nolist,nohyperlinks]{acronym}
\usepackage{float}
\usepackage{svg}
\usepackage{multicol}
\usepackage{booktabs}
\usepackage{adjustbox}
\usepackage{subfigure}
\usepackage{amsmath}
\usepackage{mathtools}
\usepackage{hyperref}
\usepackage[numbers]{natbib}
\usepackage{pifont}

\usepackage[english]{babel}

\newcommand{\V}[1]{\boldsymbol{\mathbf{#1}}}

\newcommand\BibTeX{{\rmfamily B\kern-.05em \textsc{i\kern-.025em b}\kern-.08em
T\kern-.1667em\lower.7ex\hbox{E}\kern-.125emX}}

\newif\ifImageAtTheEnd
\ImageAtTheEndfalse

\begin{document}

\runninghead{Lorentz et al.}
\title{Heavy Metal Acrobatics: Learning Dynamic Control for Skid-Steer Loaders}

\author{%
Pierre-Jean Lorentz\affilnum{1,2},
Pascal Egli\affilnum{1,3},
Edo Jelavic\affilnum{1,3},
Cliff Li\affilnum{1},
Ramitha Sundar\affilnum{2},
Harikrishnan Suresh\affilnum{2},
Paul Montgomery\affilnum{2},
Lee Redden\affilnum{2},
Matthias Müller\affilnum{4},
Ozan Sener\affilnum{4},
Vladlen Koltun\affilnum{4},
Marco Hutter\affilnum{1}}

\affiliation{%
\affilnum{1}Robotic Systems Lab (RSL), ETH Zürich, Zürich 8092, Switzerland\\
\affilnum{2}Blue River Technology (BRT), Santa Clara, CA 95054, USA\\
\affilnum{3}Gravis Robotics AG, Opfikon 8152, Switzerland\\
\affilnum{4}Intelligent Systems Lab, Intel, Jackson, WY 83001, USA}

\corrauth{Pierre-Jean Lorentz}

\email{pierrejean.lorentz@gmail.com}

\begin{abstract}
Automation of heavy machinery offers the opportunity to improve productivity in mechanized industries such as construction, mining, and material handling. Yet, automation of such machines is challenged by their complex dynamics, in part because of hydraulic actuation that involves fluids, flexible elements, and nonlinear delays. A concomitant challenge is the size and weight of these machines, which can make experiments on the physical systems exceedingly dangerous.
Here, we show that \ac{RL} in simulation, followed by transfer of the learned policies from simulation to reality, can successfully automate acrobatic maneuvers that push a heavy construction machine to the limits of its actuation and have only previously been accomplished by a small number of elite human operators in official world-record performances. We trained policies that perform autonomous acrobatic maneuvers on a John Deere 332G, an operational 5-tonne skid-steer loader. The policies can autonomously swing the machine up onto two wheels and, while balancing on two wheels, drive over rough terrain, trace out specified trajectories, spin around, and swing the bucket back and forth before landing it safely back onto four wheels. This work demonstrates that \ac{RL} and sim-to-real transfer enable dynamic autonomous control of heavy machinery, directly contributing to greater automation in heavy equipment sectors.
\textbf{Videos}: \url{https://sites.google.com/leggedrobotics.com/heavy-metal-acrobatics}

\vspace{1cm}
\textit{This is the author's version of the work. It is posted here by permission of the AAAS for personal use, not for redistribution. The definitive version was published in Science Robotics on 11 November 2026, DOI: \href{https://doi.org/10.1126/scirobotics.aef0894}{10.1126/scirobotics.aef0894}.}
\end{abstract}

\begin{acronym}
\acro{AHRS}{attitude and heading reference system}
\acro{IMU}{inertial measurement unit}
\acro{AI}{artificial intelligence}
\acro{RL}{Reinforcement Learning}
\acro{NN}{neural network}
\acro{PPO}{Proximal Policy Optimization}
\acro{MLP}{multi-layer perceptron}
\acro{DoF}{degrees of freedom}
\acro{MPC}{Model Predictive Control}
\acro{SMB}{Super Mega Bot}
\acro{SSL}{skid-steer loader}
\acro{MDP}{Markov decision process}
\acro{EKF}{extended Kalman filter}
\acro{CAN}{Controller Area Network}
\acro{CoM}{center of mass}
\acro{PWM}{pulse-width modulation}
\acro{GAE}{generalized advantage estimation}
\acro{DR}{domain randomization}
\acro{LiDAR}{Light Detection and Ranging}
\acro{CAD}{computer-aided design}
\end{acronym}

\maketitle

\begin{figure*}
    \centering
    \includegraphics[width=\linewidth]{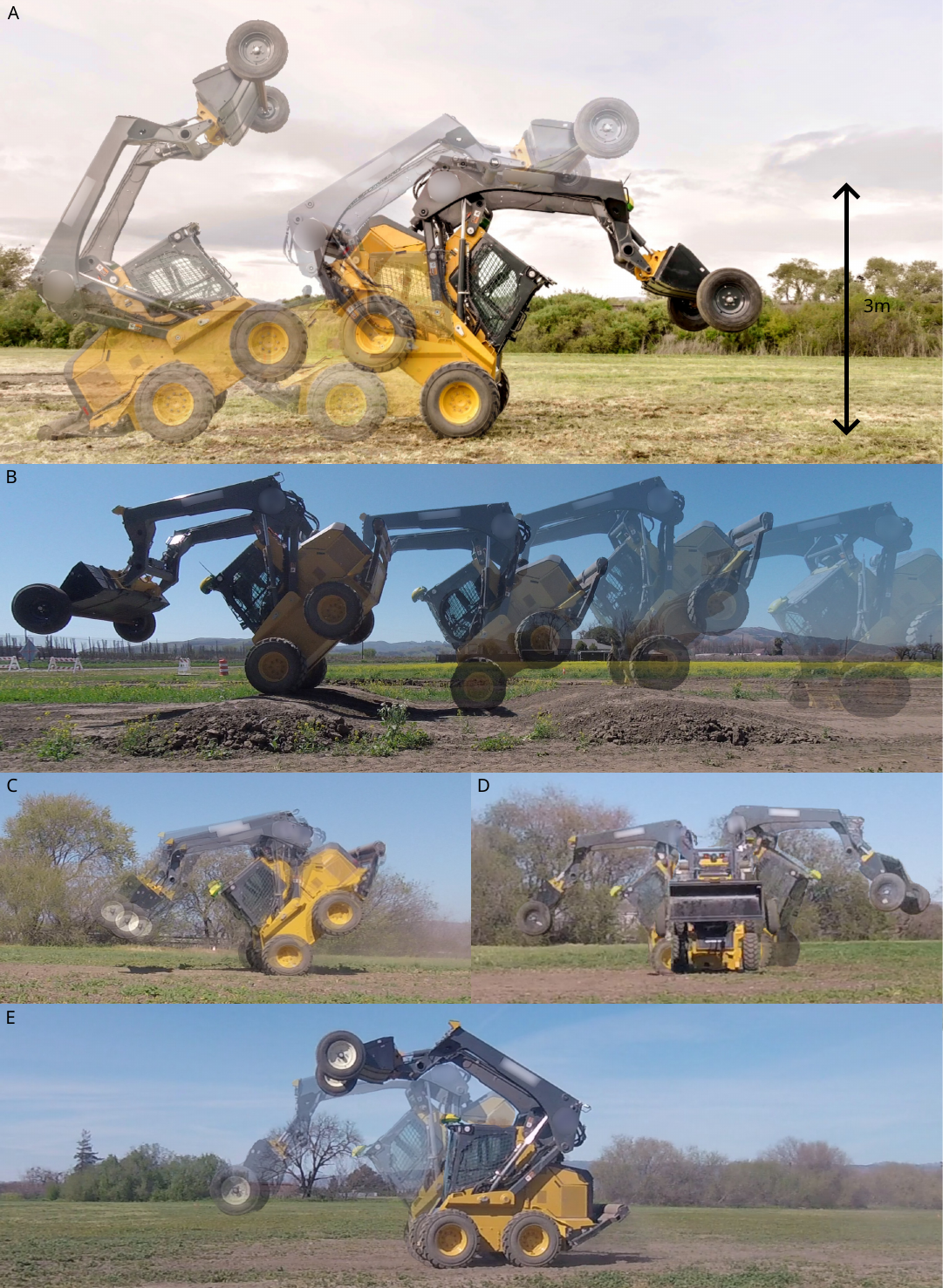}
    \caption{\textbf{Acrobatic maneuvers trained with \ac{RL} in simulation and deployed on a 5-tonne \acs{SSL}}.
        \textbf{(A)} Swing-up from four onto two wheels.
        \textbf{(B)} Driving over rough terrain on two wheels.
        \textbf{(C)} Stable balancing while moving the arm.
        \textbf{(D)} Spinning in place on the front wheels.
        \textbf{(E)} Soft landing after balancing on two wheels.
    }
    \label{fig:title}
\end{figure*}

\section*{Introduction}

Heavy machinery presents an opportunity for automation, not only to increase productivity and save costs but also to alleviate workforce shortages and increase safety~\cite{mckinsey_the_next_normal}.
Analytical models of the dynamics of such machines are often highly inaccurate because of hydraulic elements that involve fluids and flexible conduits; highly nonlinear actuator dynamics with substantial delays; and manufacturing tolerances; joint play; and friction effects.
Model-based control approaches such as \ac{MPC} have demonstrated strong performance in hydraulic systems, including trajectory tracking~\cite{ma2024datadrivenmultistepnonlinearmodel, paz2025fulldynamicsrealtimenonlinearmodel}, vibration control~\cite{dynamic_improvement}, and coordinated motion control~\cite{kalmari_forestry}. However, highly dynamic behaviors involving underactuated balancing and frequent contact changes introduce additional challenges related to model accuracy, contact handling, and real-time optimization.

A promising complementary path is \acf{RL}, which has facilitated breakthroughs in legged locomotion~\cite{hwangbo2019learning,lee2020learning, Miki2022sciro,Hoeller2024:SciRob} and agile flight~\cite{scaramuzza_nature,Song2023:SciRob}.
\Ac{RL} policies excel at concurrently managing a multitude of inputs and outputs, making them exceptionally well suited for the control of intricate robotic systems with many degrees of freedom and frequent contact changes. The basic approach is based on sim-to-real transfer: The policy is trained in simulation and is subsequently deployed in the physical world.

Recent work has applied \ac{RL} and sim-to-real transfer to statically stable tasks such as breaking rocks with an impact hammer~\cite{Samtani2023_rock_breaking}, soil-adaptive excavation~\cite{Egli2022}, driving a heavy forestry vehicle with active
suspension over rough terrain~\cite{Wiberg2023_forestry_driving}, and controlling hydraulic excavator arms for surface grading~\cite{Egli2021}.

In this Research Article, we demonstrate that \ac{RL}, combined with sim-to-real transfer, can enable the autonomous execution of highly dynamic and agile maneuvers with heavy machinery. Specifically, we trained a 5-tonne \acf{SSL} to perform a series of acrobatic behaviors. We taught the machine to swing up from a stable four-wheel stance onto its two front wheels by actively accumulating momentum through coordinated back-and-forth motion. Once balanced, it can track arbitrary velocity commands to follow specified trajectories, traverse rough terrain, and even perform controlled spins, all while maintaining the two-wheel balancing configuration. Moreover, the learned policies enable the machine to remain stable while the bucket and arm are in motion and to smoothly transition back to a four-wheel stance.
These acrobatic behaviors operate the machine at the limits of its available actuation, involving full-speed motion and maximum control inputs. They involve maneuvers that are typically mastered only by elite human operators. Notably, several of the behaviors -- balancing on two wheels, driving while balancing (wheelie), and spinning in place while balancing -- mirror those featured in official world-record performances~\cite{guinnessSkidsteerBalancing, guinnessSkidsteerSpinning, guinnessSkidsteerWheelie}. Because of the weight of the machine and the highly dynamic nature of the maneuvers, mistakes in execution risk substantial damage to the equipment and its surroundings.
Beyond the artistic nature of the presented maneuvers, the learned behaviors may also be relevant in practice. Controlled balancing of an \ac{SSL} could, for example, enable autonomous self-loading onto trucks without ramps or reduce tire scrubbing and ground damage during maneuvering on sensitive terrain. More broadly, the presented behaviors demonstrate dynamic whole-body control of heavy hydraulic machinery, which may facilitate future applications involving obstacle negotiation, recovery maneuvers, or operation in highly unstructured environments.
The different maneuvers are illustrated in \autoref{fig:title} and captured in Movie~1.

\section*{Results}

We performed experiments on a John Deere 332G, an \ac{SSL} with four wheels on two fixed axles, a hydraulic drive train, and an operating weight of 5 tonnes.
When operated by a human, the machine is driven via two steer-by-wire joysticks, one commanding the linear motion and the other the angular speed. The machine's low-level controllers convert the joystick inputs to different speed commands for the left- and right-side wheels. Steering is achieved through skidding by supplying left and right wheels with different commands. The machine also has a bucket that can be raised above the cabin to move material.

\begin{figure}
    \centering
    \includegraphics[width=1\linewidth]{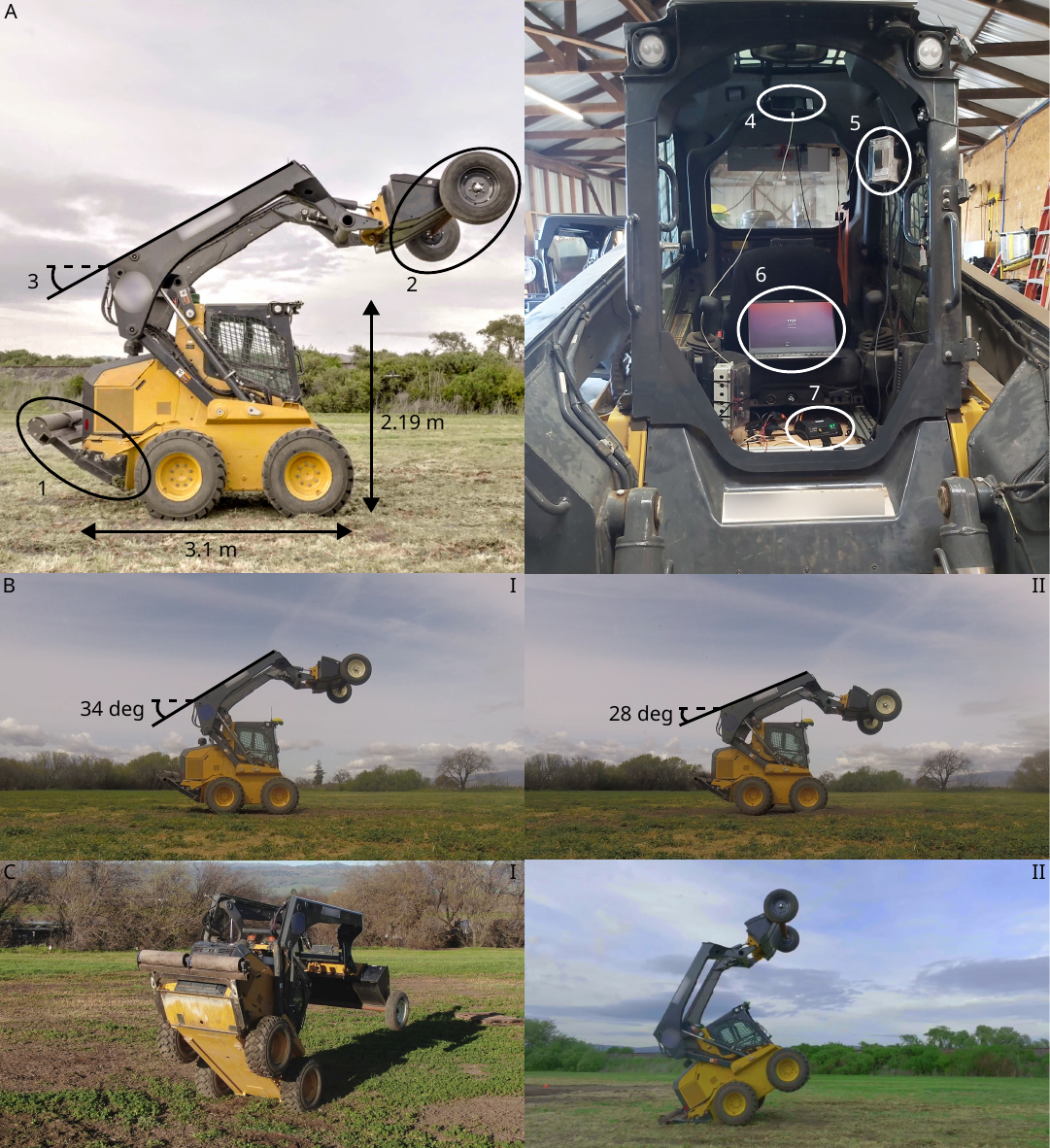}
    \caption{\textbf{Hardware description.} \textbf{(A)} John Deere 332G \ac{SSL}. For safety, a wheelie bar with a rolling tube was mounted at the back (1) and wheels were mounted on the bucket (2). Two \acp{IMU} were added: one on the chassis (not visible) and another one on the arm to monitor its angle (3). Inside the machine, two internet devices (4) and (7) ensured wireless connectivity for the laptop (6), which ran the controller and interfaced the machine's \ac{CAN} bus. A physical switch (5) allowed us to enable autonomous mode only when needed.
    \textbf{(B)} Different positions of the arm. (I) is the upper joint limit ($34^\circ$), and (II) is the lowest position possible to maintain balance without touching the ground ($28^\circ$).
    \textbf{(C)} Underside view of the wheelie bar (I) and side view during swing-up (II). Note that the wheelie bar does not change the machine's ground clearance and remains unloaded during swing-up as it is bent upward.
    }
    \label{fig:hardware_description}
\end{figure}

For our experiments, we intercepted the joystick signals and used a computer to provide control commands. Wheel encoders are installed ex-works, and their measurements could be accessed via a \acf{CAN} bus. In addition, two \acfp{IMU} were retrofitted: one on the machine's base and one on the arm to monitor the arm configuration. A standard Linux laptop [Intel Core i7 central processing unit (CPU), 16~GB RAM (random-access memory), no graphics
processing unit (GPU)] was used to interface the machine's \ac{CAN} bus and execute the control policies.
To protect the machine from being damaged during testing, a bar with a rolling tube, the wheelie bar, was attached to its back to prevent the machine from tipping over backward. Note that it did not change the machine's ground clearance and remained unloaded during swing-up as it was bent upward. Furthermore, wheels were added on both sides of the bucket to prevent the arm from being damaged if it tipped forward and the shovel contacted the ground. The machine and the modifications required for autonomous operation are illustrated in \autoref{fig:hardware_description}.

We trained controllers for different tasks, fully in simulation, without expert demonstrations. Starting without any prior machine-specific knowledge or a preexisting model of the machine, we present a practical procedure to characterize the actuation dynamics and identify the most important model parameters needed to bridge the sim-to-real gap such that policies can be directly deployed on the real machine.

\subsection*{Swing-Up and Stabilize}
\begin{figure*}
    \centering
    \includegraphics[width=0.85\linewidth]{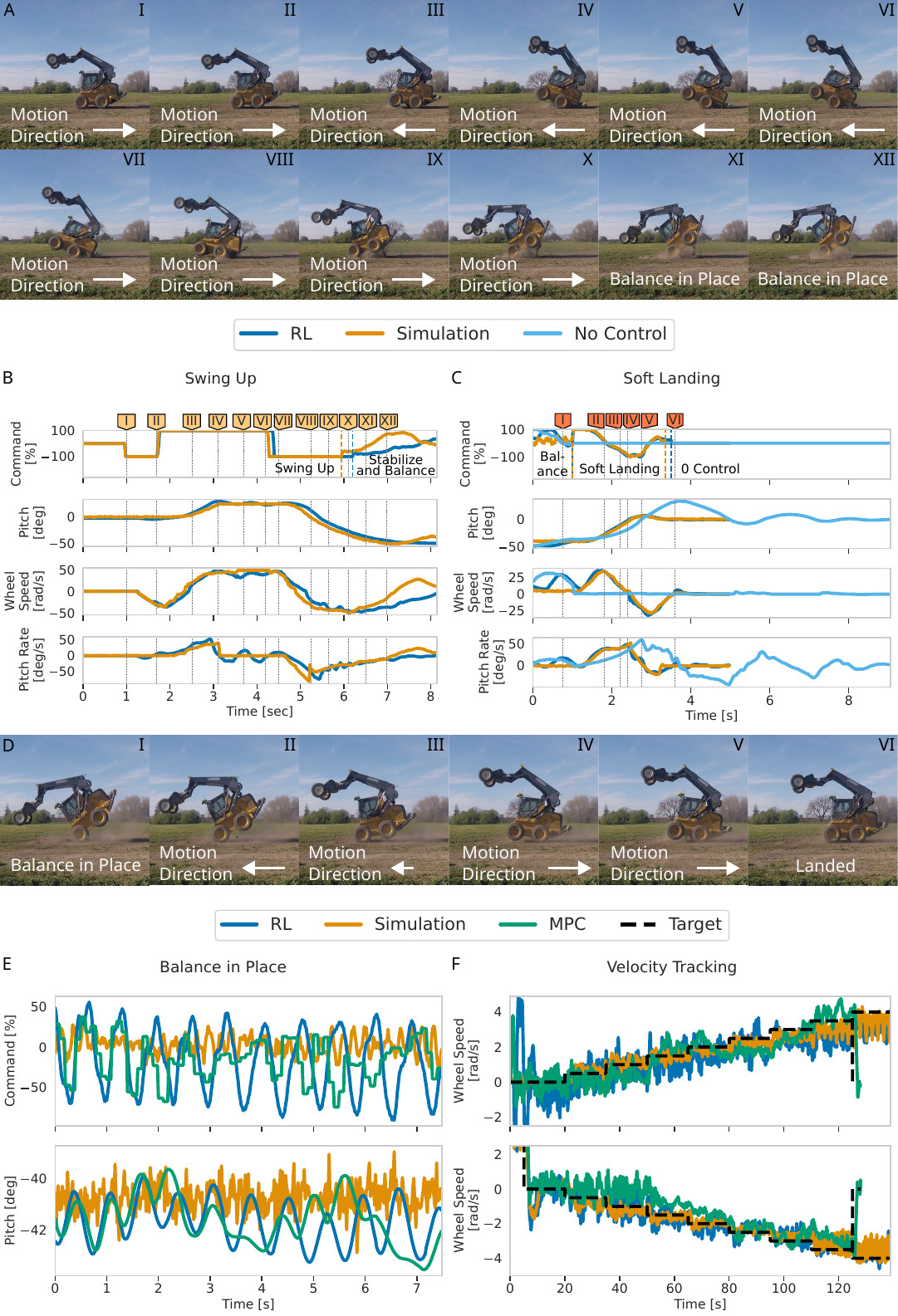}
    \caption{\textbf{Hardware deployment.}
    \textbf{(A)} Image sequence of the swing-up maneuver.
    \textbf{(B)} Swing-up in simulation and reality. First, the swing-up controller is executed. When the machine's pitch reaches $-30\si{\degree}$ (dashed lines), the stabilize-and-balance controller is activated. The roman numerals refer to the corresponding images of (A).
    \textbf{(C)} Soft-landing sequence in simulation and on the real \ac{SSL} compared with zero command. In the beginning, the balancing controller is active. The landing controller is triggered at $t=1~\si{\second}$ and is active after that. When the pitch, pitch rate, and wheel speed are close to zero, the commands are set to zero. If the soft-landing controller is inactive and the action is set to zero immediately after balancing, the \ac{SSL} falls onto four wheels. However, the front wheels rebound, causing the machine to rock back and forth.
    \textbf{(D)} Image sequence of the soft-landing maneuver with the roman numerals corresponding to the time stamps in (C).
    \textbf{(E)} Balancing in place with \ac{RL} in simulation and on the real machine, with the \ac{MPC} baseline.
    \textbf{(F)} Velocity tracking with \ac{RL} and \ac{MPC}. The target wheel velocity is increased/decreased gradually up to the maximal/minimal velocity that each controller can successfully track without touching the ground with the bucket or the rear underside. Positive target velocity tracking starts with the swing-up maneuver. Negative target velocity tracking starts with the transient behavior of switching from a positive target to a zero-velocity target.
    }

    \label{fig:figure_1}
\end{figure*}

The Swing-Up and Stabilize maneuver started with the machine at rest with all four wheels on the ground and ended with it balancing on its two front wheels (\autoref{fig:figure_1}A). To achieve this behavior, two separate controllers were trained: The Swing-Up controller brought the \ac{SSL} from four to two wheels by accumulating momentum through forth and backward motions, and the Stabilize-and-Balance controller stabilized the machine in the balancing configuration.
The transition between these controllers was triggered when the machine's pitch angle reached -30$^\circ$.
\autoref{fig:figure_1}B shows the controller actions and the state of the machine during a swing-up maneuver with the arm in the nominal configuration in reality and in simulation. The swing-up controller exhibited a bang-bang control behavior. First, the policy commanded the machine backward at full speed, then forward to lift the front wheels, and then lastly backward at full speed to swing the machine up onto the front wheels. The stabilize-and-balance controller slowed down the swing-up motion and regulated the machine in the balancing state. Overall, the simulation and reality agreed well, and the controller was robust to the residual sim-to-real gap.

To assess robustness, we ran the controllers with different arm configurations multiple times within the feasible limits, where the upper bound was given by the joint constraints and the lower bound was given by the minimum to balance without ground contact. The policy did not observe the arm configuration, which was randomized between $29^\circ$ and $31^\circ$ during training. We considered a swing-up successful if the machine balanced on the front wheels and a failure if it fell onto the bucket or back onto the rear wheels. The results are summarized in \autoref{tab1:swingUpProba}. As in simulation, the controllers reliably performed the swing-up maneuver on the real machine when the arm was in a configuration experienced during training. However, when the arm was higher than during training, the policy overshot and the bucket touched the ground in 3 of 14 trials. We attribute this to the increased inertia of the system when the arm was higher, which makes it harder to slow down after a swing-up and avoid touching the ground with the bucket.

\begin{table}
\small\sf\centering
\caption{Performance evaluation of the swing-up maneuver. The feasible range of arm angles corresponds to $[28^\circ, 34^\circ]$, where the upper limit is given by the joint constraints and the lower limit by the minimum to balance without touching the ground. During training, the arm configuration was randomized between 29$^\circ$ and 31$^\circ$.}
\label{tab1:swingUpProba}%
\begin{adjustbox}{max width=\columnwidth}
\begin{tabular}{@{}llllll@{}} \toprule
\textbf{Arm} & \textbf{Trials} & \textbf{Success} & \textbf{Failure}  & \textbf{Failure on}  & \textbf{Success}  \\
 \textbf{Angle}& & &  \textbf{on Bucket} & \textbf{4 Wheels} & \textbf{Prob.} \\
 \midrule
28$^\circ$   & 11   & 11  & 0 & 0 & 100\%\\
31$^\circ$  & 12   & 12  & 0 & 0 & 100\%  \\
34$^\circ$   & 14   & 11  & 3 & 0 & 78.6\%\\
 \bottomrule
\end{tabular}
\end{adjustbox}
\end{table}

\subsection*{Soft Landing}

The soft landing controller brought the machine from balancing on the front wheels to a four-wheeled stance, focusing on minimizing impact forces. We compared the performance of the learned maneuver to simply sending no commands. The controller's performance on the real machine, along with snapshots of the maneuver, is shown in \autoref{fig:figure_1} (C and D).
As soon as the controller was activated [time step (t) = 1~\si{\second}], the action increased to 100\%, corresponding to full speed forward. This ensured that the machine fell back onto the rear wheels as fast as possible. Then, the action decreased until it reached $-100\%$ (t = 2.5~\si{\second}), corresponding to full speed backward. Driving backward minimized the impact forces and prevented the front wheels from bouncing off the ground after initial contact. The \ac{RL} controller could softly land the machine back on four wheels in simulation and in reality, whereas sending no commands resulted in a severe impact that caused the front wheels to bounce back off the ground, making the machine rock back and forth, as shown in Movie S1.

\subsection*{Balance and Track Velocities}

The goal of the balance-and-track-velocities maneuver was to drive and steer the machine while balancing on the front wheels.

In the first experiment, we compared the performance of balancing in place of the \ac{RL} controller to the \ac{MPC} baseline (see the \ac{SSL} Model Predictive Control Baseline in \ref{supp:MPC} of the Supplementary Materials). The results are shown in \autoref{fig:figure_1}E. Both controllers could successfully balance the \ac{SSL}. The \ac{RL} controller balanced the machine with $\pm1.6$$^\circ$ oscillations around the nominal position, on par with the \ac{MPC} controller.
Compared with the \ac{RL} controller in simulation, the real machine showed a $\sim$2$^\circ$ smaller average balancing pitch angle caused by the remaining sim-to-real gap. Nevertheless, the controller was robust to this discrepancy.

In a second set of experiments, we analyzed the tracking performance of the controllers. The results are shown in \autoref{fig:figure_1}F. We increased the velocity target stepwise from $0$ to $4~\si{\radian\per\second}$, the maximum value during policy training for the \ac{RL} controller. The \ac{MPC} baseline failed: It fell forward onto the bucket or backward onto the rear wheels for velocities beyond 3.5~\si{\radian\per\second}.
When tracking positive targets, the high velocities between $0$ and $10~\si{\second}$ on \autoref{fig:figure_1}F originated from the swing-up maneuver. When tracking negative targets, \autoref{fig:figure_1}F started with the transient behavior of the controllers, switching from tracking their maximum positive target to tracking a $0$ target. That transient part at the start (first 20~\si{\second}) shows that the controllers were able to handle abrupt changes of the velocity target. The full experiment is shown in Movie S2.

\begin{figure*}
    \centering
    \includegraphics[width=1\linewidth]{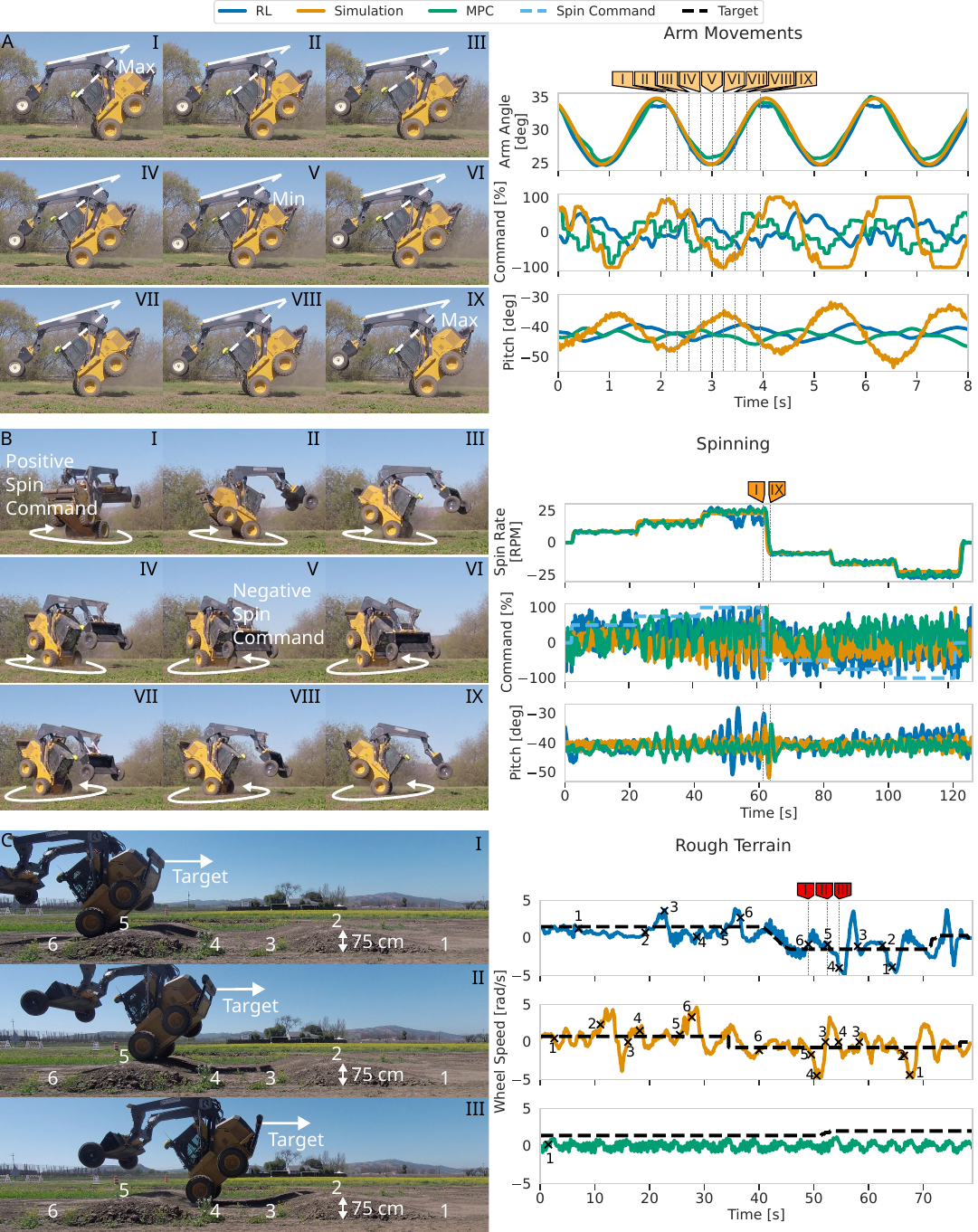}
    \caption{\textbf{Robustness analysis.} For all panels, the corresponding snapshot times are indicated in the plot with roman numerals.
    \textbf{(A)} Robustness to arm movements of the \ac{RL} controller and the \ac{MPC} baseline. The arm is moved around its nominal position. Between (I) and (V), the arm moves downward; between (V) and (IX), the arm moves upward.
    \textbf{(B)} Robustness to spinning at different wheel speeds of the \ac{RL} policy and the \ac{MPC}. Spinning is achieved by overlaying a machine rotation command on the policies' actions.
    \textbf{(C)} Traversing rough terrain. The \ac{SSL} successfully navigates the $75~\si{\cm}$ bumps, driving forward and backward. The image sequence shows the \ac{SSL} driving backward over the bumps. The green curve shows the \ac{MPC} not being able to traverse the first bump, even when increasing the target velocity.
    }
    \label{fig:figure_2}
\end{figure*}

Next, we evaluated the robustness of the \ac{RL} controller to arm movements, both in simulation and on the real system, and compared its performance with the \ac{MPC} baseline. Moving the arm shifts the center-of-mass position, thereby altering the balancing angle and generating inertial forces.
The results are shown in \autoref{fig:figure_2}A. In simulation, the \ac{RL} controller responded strongly to the disturbance but remained stable. The machine's pitch-angle oscillations initially increased slightly before converging to a constant amplitude. When deployed on the real machine, both the \ac{MPC} and \ac{RL} controllers achieved comparable performance and maintained balance despite the perturbation. The oscillation amplitudes were lower than in simulation, which we attribute to the remaining sim-to-real gap. Movie S3 shows the full experiment.

We also tested the policy's robustness by overlaying rotational yaw commands over the policy output, causing the machine to spin. The effect of spinning on the controller was evaluated at three different rotational speeds, in each direction: $\pm50~\si{\percent}, \pm75~\si{\percent}, \pm100~\si{\percent}$. Note that the rotational speed is a direct input to the machine provided by the left joystick during manual operation. This led to differences in wheel speed between the left and right wheels of 3, 5.7, and 8.4~\si{\radian\per\second}, respectively.
The results are shown in \autoref{fig:figure_2}B. The \ac{RL} controller was robust to the spinning commands. When spinning at the maximum velocities, the controller stayed balanced but oscillated more than at lower angular velocities. A similar behavior could be observed in simulation, where the policy consistently balanced the skid steer with spin commands up to $\pm75~\si{\percent}$ and oscillated at $\pm100~\si{\percent}$. In comparison, the \ac{MPC} controller managed to balance the machine stably, even at the maximal velocities, but showed larger oscillations at $+75~\si{\percent}$ spinning. This experiment is shown in Movie S4.

We also steered the \ac{SSL} over nonflat terrain (\autoref{fig:figure_2}C) with 75~\si{\cm}-high undulations. Note that the controller was only trained on flat terrain and was unaware of obstacles. Nonetheless, it successfully kept the machine balanced and could drive over bumps both forward and backward.
\autoref{fig:figure_2}C shows the wheel speed during the maneuver. When the \ac{SSL} reached a bump, for example, at point 1 in \autoref{fig:figure_2}C (blue curve), the machine slowed down and then accelerated again. Similarly, when going down, for instance, at point 2 in \autoref{fig:figure_2}C (blue curve), it accelerated because of gravity and then the policy reacted by slowing down the desired wheel speed. \autoref{fig:figure_2}C (orange curve) shows the corresponding maneuver in simulation.
The \ac{MPC}, shown in \autoref{fig:figure_2}C (green curve), was not able to handle this terrain. It stayed balanced but was unable to traverse the undulations in the terrain, even when the target velocity was increased. This is shown in Movie S5.

\section*{Discussion}
We successfully trained \ac{RL} policies to perform highly dynamic maneuvers on a full-size \ac{SSL} with a hydraulic drivetrain. The presented approach does not rely on any prior machine-specific knowledge or a preexisting model of the machine but instead provides a practical procedure for characterizing the actuation dynamics and identifying the most important model parameters to bridge the sim-to-real gap. With this, domain randomization techniques suffice to train \ac{RL} controllers that can be directly deployed on the real hardware. We deliberately trained individual skills to keep the tuning effort and number of parameters low compared with learning the full maneuver, or components, in a single policy.

Compared to a state-of-the-art \ac{MPC} baseline, the presented \ac{RL} approach exhibits similar performance in simpler tasks such as balancing and velocity tracking.
The \ac{MPC} controller exhibits milder oscillations at higher spinning rates,
whereas the \ac{RL} policy can track higher velocities without losing balance. Furthermore, the \ac{RL} policy generalizes to nonflat terrains, which stymie the \ac{MPC} controller. We attribute this robustness to domain randomization during training, whereas the \ac{MPC} controller without task-specific extensions struggles to compensate for such terrain disturbances even with increased commanded wheel velocities.
More complex tasks that require longer planning horizons and multiple contact switches, such as the swing-up, present a challenge for \ac{MPC} because of the nonconvexity of the problem.
Multicontact scenarios necessitate dedicated planners to initialize the optimization \cite{jelavic2023lstp}, which requires manually extending the model with additional contact points specific to a particular task.
In contrast, with the \ac{RL} approach this can be achieved by simply changing the reward definition and initialization procedure while leaving the overall training procedure unchanged. Furthermore, the computational burden is shifted to simulation during training and results in relatively small neural networks, whose inference at runtime is almost free compared with that of optimization- or sampling-based approaches.
\autoref{tab:rl_vs_mcp_overview} summarizes the performance of the \ac{RL} controller compared with the \ac{MPC} baseline.

\ifImageAtTheEnd \else
\begin{table}[h]
\caption{Summary of the performance comparison between \ac{RL} and the \ac{MPC} baseline. \ding{51} indicates ability to perform the task, and \ding{55} indicates failure to perform the task.}
\label{tab:rl_vs_mcp_overview}%
\begin{adjustbox}{max width=\columnwidth}
\begin{tabular}{@{}lcc@{}} \hline
\textbf{Maneuver} & \textbf{RL} & \textbf{MPC} \\ \hline
Swing-up & \ding{52} & \ding{55}\\
Static balancing & \ding{51}  & \ding{51}  \\
Velocity tracking &
\ding{52} \textbf{max} $\boldsymbol{\pm 4\ \si{\radian\per\second}}$ &
\ding{51} max $\pm 3.5\ \si{\radian\per\second}$ \\

Spinning & \ding{51}  & \ding{51} \\

Arm movements & \ding{51}  & \ding{51}\\

Nonflat terrain & \ding{52} & \ding{55}\\

Soft landing & \ding{52} & \ding{55}\\\hline

\end{tabular}
\end{adjustbox}
\end{table}
\fi

With this work, we showcase a practical approach for acquiring dynamic control skills with \ac{RL} on full-scale hydraulic machinery, which can serve as building blocks for more complex autonomous tasks in mechanized industries such as construction, mining, agriculture, and other heavy equipment sectors. We believe that training individual control skills is a prerequisite to achieving more complex and longer-horizon tasks. For simplicity, we composed the overall behavior using hand-crafted switching conditions. This works robustly and reliably because the state distributions are overlapping between skills; for instance, the stabilize-and-balance controller is initialized in configurations reached by the swing-up controller, as outlined in more detail in the following sections. For more complex tasks requiring broader contextual reasoning, such as obstacle negotiation or self-loading onto a truck without ramps, the switching logic could be replaced by a higher-level decision-making mechanism. A viable path is hierarchical \ac{RL}, where a high-level policy is trained to select and activate the most appropriate low-level skill~\cite{Hoeller2024:SciRob}. Although the use of computationally lightweight low-level skills, as presented in this work, is beneficial for reducing training complexity and computation time, incorporating model-based controllers into the high-level training loop is also possible, albeit at a much higher computational cost~\cite{sciro_deep_tracking_control}.

Future directions include extending the capabilities of the \ac{RL} controllers by using sensor modalities such as \ac{LiDAR} or cameras to perceive the environment and learn to traverse challenging terrain~\cite{Miki2022sciro}. Furthermore, the individual skills could be distilled into a single policy. This may improve performance and generalization and should be investigated in the future.
Last, privileged learning \cite{lee2020learning} or online adaptation \cite{Yu2020} can be explored to further advance high-performance autonomous control of heavy machinery.

\section*{Materials and Methods}

\subsection*{Overview}
Our approach is based on \ac{RL} in simulation and is illustrated in \autoref{fig:figure_3}A. It comprises the following steps: First, we created a rigid-body model of the machine accounting for both kinematic and dynamic properties of the machine. Kinematic properties such as wheel diameter, wheelbase, position of the machine's rear underside, and the position of the bucket were measured directly on the machine. Dynamic properties such as the mass and the center-of-mass location were measured with truck scales and derived on the basis of the equilibrium of moments. Second, we created a model of the actuation. Therefore, we collected real-world data while moving each degree of freedom of the machine with randomized commands, thus capturing the dynamics of the motor, and trained a machine-learning model to predict the actuator behavior. Third, we formulated an appropriate reward structure for \ac{RL} training of the policy in simulation. The simulation consists of the rigid-body model and the actuator model. Last, we deployed the policy on the physical hardware when it performed as required in simulation.

\subsection*{Simulation}
\subsubsection*{Actuator Model}

A critical element of the simulation is the actuation dynamics.
Hydraulic actuators exhibit strong nonlinearities and delays, which are very laborious to model from first principles and specific to every machine. As shown by Hwangbo et al.~\cite{hwangbo2019learning}, even sophisticated analytical models fail to capture the dynamics of series-elastic drives accurately enough for successful sim-to-real transfer. Therefore, we collected data on the real machine to train a general-purpose neural network that accurately captures the actuation behavior, including delays and nonlinearities~\cite{hwangbo2019learning,towards_rl_based}.

For the hydraulic actuators on the \ac{SSL}, control inputs correspond to wheel velocities. This was demonstrated in a simple experiment in which we ran the same control input sequence twice, once with the machine moving on flat ground and once with the machine lifted off the ground such that the wheels could rotate freely. The results are shown in \autoref{fig:figure_3}B. The measured wheel speed was almost identical in both cases. Therefore, the actuator model was trained to map a history of control inputs and wheel speeds up to time step $t$ to the wheel velocity at $t+0.02~\si{\second}$. The history is needed to capture hysteresis effects and delays.

An initial dataset was collected by moving the machine back and forth with different command sequences. We used step, sinusoidal, ramp, and constant zero inputs, with amplitudes up to 100\% of the maximum command, for a total of 1.5~\si{\hour}. An excerpt of this process is shown in Movie~1. With these data, a successful balancing controller could be trained. The dataset was then augmented with 20~\si{\min} of data collected during balancing, which slightly improved the performance of the balancing controller. The exact composition of the dataset is outlined in \ref{tab:ssl_actuator_model_data_collection}. The history length and sparsity were chosen empirically to maximize the accuracy of the actuator model's predictions. Specifically, we covered $1.8~\si{\second}$ of history with $30$ data points (50~\si{\hertz} measurement rate and stride 3). The actuator model architecture is a neural network with $3 \times 32$ fully-connected layers with SoftSign activation and a linear output layer. It was trained using the Mean Squared Error loss and Adam Optimizer~\cite{adam_opt} until the validation loss stopped improving on a single-workstation PC equipped with an Nvidia GeForce RTX 2070 Super GPU and an AMD Ryzen 5 2600 CPU.
The actuator model was then evaluated on two different datasets: one collected while the machine was balancing and one collected while driving the machine back and forth manually, with all four wheels on the ground. \ref{tab:ssl_act_model_eval} summarizes the quantitative evaluation of the learned actuator model.

The torques applied to the wheels in simulation were generated by a Proportional Derivative (PD) controller that takes the velocity prediction of the actuator model as reference input (\autoref{fig:figure_3}A). \autoref{fig:figure_3}C compares the wheel speed on the real machine to the simulation for the same command sequences, showing the close alignment between simulation and deployment on the physical machine.

\subsubsection*{Rigid-Body Model}
\begin{figure*}
    \centering
    \includegraphics[width=1\linewidth]{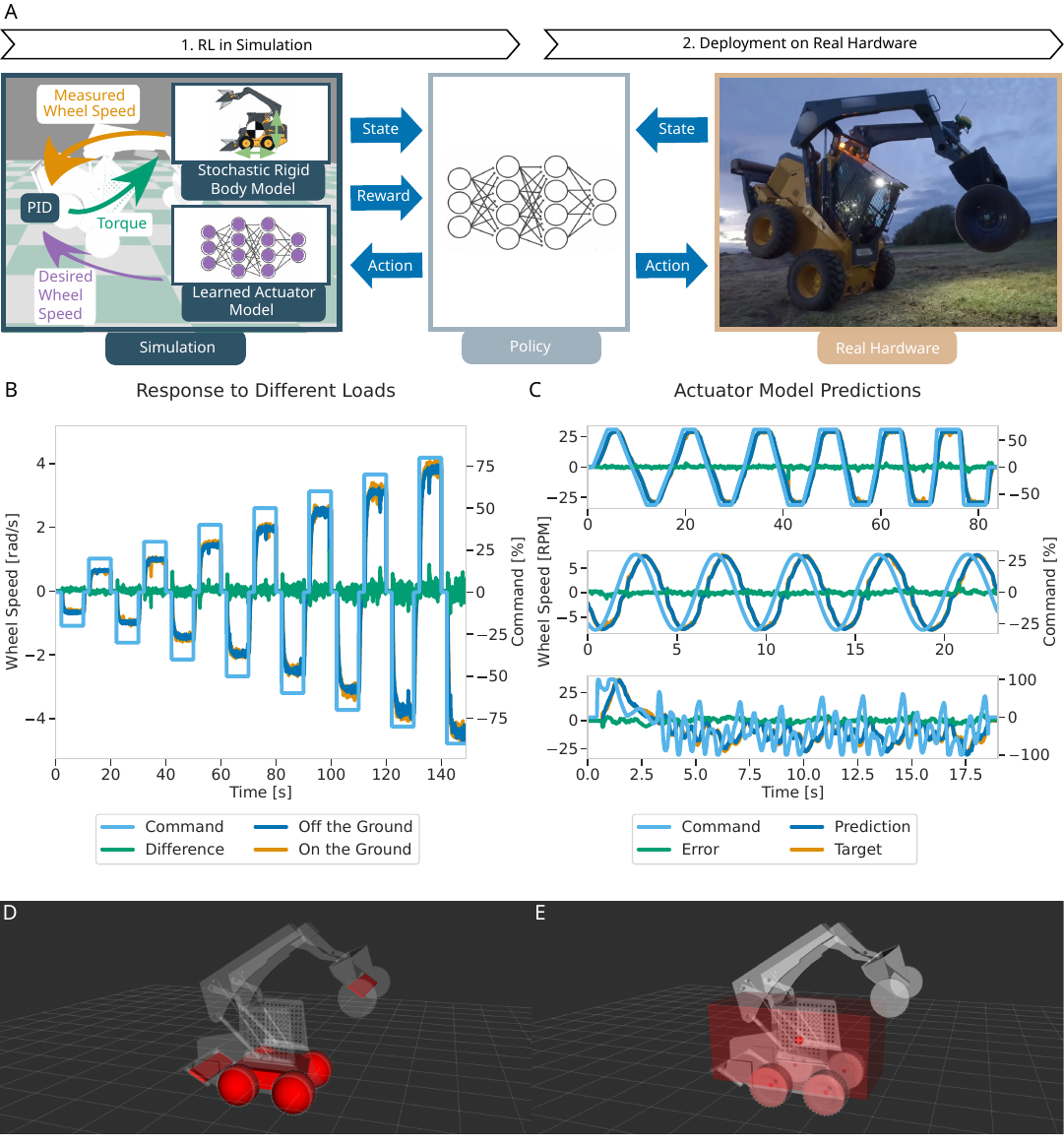}
    \caption{\textbf{Method overview.} \textbf{(A)} The controllers were first trained in simulation and then deployed on real hardware. The simulation incorporated an actuator model trained to predict wheel speeds on the basis of real-world data and a rigid-body model whose properties were randomized during training around the measured values.
    \textbf{(B)} Wheel speed for the same joystick input, rotating freely in the air (machine is lifted with a car jack) and when placed on the ground (thus moving the entire \ac{SSL}).
    \textbf{(C)} Comparison of the wheel speed in simulation and on the real machine for different types of signals: ramps, sinusoidal, and commands from the controller while balancing the machine.
    \textbf{(D)} Hand-measured collision geometry of the rigid-body model (highlighted in red). Generic mesh for visualization only in gray.
    \textbf{(E)} Visualization of the wheel and machine centers of mass (red dots) and the inertia approximation with a uniform mass distribution over the red cuboid for the cabin and four cylinders for the wheels. Note that the machine's center of mass changes depending on the arm configuration.
    }
    \label{fig:figure_3}
\end{figure*}

In addition to the actuator dynamics, simulating the system required rigid-body properties such as collision geometry and mass distribution.
Because no accurate \ac{SSL} machine model was available from the manufacturer, the shape of the robot was approximated as a cuboid main body with four cylindrical wheels with uniform density and an additional flat box for the machine's rear underside. Note that the wheelie bar and the bucket wheels served purely as safety devices and were excluded from the collision model to prevent the agent from exploiting them during training.
The kinematic and dynamic properties were determined by simple means: Link lengths were measured with a tape measure, and the mass distribution was measured with truck scales via equilibrium of moments \cite{omnicalculator}. Note that the mass of the arm is lumped into a single center of mass of the machine, which moves depending on the arm configuration. \autoref{fig:figure_3} (D and E) visualizes the contact geometry and the inertial properties, respectively.

\subsection*{Reinforcement Learning}
To train \ac{RL} policies for different maneuvers in simulation, we formulated the control problem as a discrete-time Markov decision process, where an agent interacts with the environment. The state of the environment is denoted by ${\V s_t \in \mathcal{S}}$.
At each time step $t$, the agent receives an observation $\V o_t \in \mathcal{O} \subseteq \mathcal{S}$ and takes an action consisting of motor commands
$\V a_t \in \mathcal{A}$ sampled from a stochastic Gaussian policy $\pi_{\V{\theta}}(\V a | \V o_t) \sim \mathcal{N}(\V{\mu_{\theta}}, \V{\sigma}^2)$. The mean action $\V{\mu_{\theta}}$ is the output of a neural network with parameters $\V{\theta}$, and the variance $\V{\sigma}^2$ is independent of the observation. In addition, the agent receives a scalar reward ${r_t(\V s_t, \V a_t) \in \mathcal{R}: \mathcal{S} \times \mathcal{A} \rightarrow \mathbb{R}}$. The agent's objective is to maximize the infinite-horizon return $\mathbb{E} \big [ \sum_{t=0}^{\infty}{\gamma^t r_{t}} \big ]$, where the discount factor $\gamma \in [0,1)$ trades off between current and future rewards and renders the return numerically finite.

The policies were trained using \ac{PPO} with generalized advantage estimation~\cite{stable-baselines} on the same workstation that was used to train the actuator model. Experience generation in the RaiSim simulation engine~\cite{raisim} took place on the CPU and policy updates on the GPU. The policy and critic networks each consisted of two-layered feed-forward neural networks with tanh nonlinearities and linear output layers, which received the same observation. Once trained, the policies were deployed on the physical hardware. Upon deployment, the observations, which are the controller inputs, were sourced from sensors on the physical machine.

\subsubsection*{Observations and Actions}
For all maneuvers, the observation and action spaces are the same. The agent observed the pitch angle, calculated by the internal extended Kalman filter of the \ac{IMU};
the pitch rate from the \ac{IMU};
the average wheel speed from the left- and right-side encoders;
the previous action from the policy, which is the control signal to the machine;
and the target commands for linear velocities (only for the command tracking task).

A history of these observations was also provided to the agent. The history length was adapted on the basis of the task as outlined in the following sections. Gaussian random noise, with mean $\mu = 0$ and standard deviation adapted for each task, was added to the measurements in the simulation. This made the controller more robust to real sensor imperfections and helped overcome mismatches between the simulation and the physical hardware. Controllers trained without sensor noise in simulation did not perform reliably once deployed in the physical world. The numerical values and sensor details are summarized in \ref{tab:ssl_observed_variables}. The action consisted of the linear joystick command. Actions and observations were not scaled because their magnitudes and distributions are similar.

\subsubsection*{Domain Randomization}
Because of the simplifications made during modeling, a sim-to-real gap was still present. To ensure that the trained policy was robust to this discrepancy, we used domain randomization \cite{Peng2018_DR} and slightly altered the simulation model in every training episode. At the beginning of every episode, the position of the arm, the position of the center of mass, and the inertia of the machine were sampled from a uniform distribution according to the values listed in \ref{tab:ssl_dr}.

\subsubsection*{External Disturbances}
To further robustify the policy, external disturbances were applied at the center of mass of the \ac{SSL}~\cite{Valassakis2020}. This disturbance was sampled from a uniform distribution and persisted for a sampled period. The corresponding hyperparameter settings are listed in \ref{tab:ssl_external_disturbance}.

\subsection*{Maneuvers}
We tackled the tasks of swinging the machine up onto its front wheels, keeping it balanced on two wheels, tracking desired velocities while balancing, and safely landing back onto four wheels. The entire maneuver consisted of four subtasks, each trained individually in simulation: swing-up, stabilize and balance, balance and track velocities, and soft landing. For deployment, switching between the subtasks was defined on the basis of simple rules. This was enabled by overlapping state distributions between skills. In the following, the training procedure for the four maneuvers is explained in detail. The training hyperparameters and reward coefficients for the different maneuvers are listed in \ref{tab:SSL_training} and \ref{tab:ssl_reward_function_paramters}, respectively.

\subsubsection*{Swing-Up}

The goal of the swing-up maneuver was to bring the machine from a stationary four-wheel stance onto its two front wheels with a pitch angle of $-30^\circ$ or less. This could only be achieved by rocking the machine back and forth to accumulate energy such that, eventually, the machine could be brought onto two wheels.

\paragraph{Episode Initialization}
The simulation always started with the \ac{SSL} at rest on four wheels (\autoref{fig:figure_4}A).

\paragraph{Reward}
The reward was defined as a function of the machine's pitch angle as
\begin{equation}
\begin{split}
    \label{eq:swing_up_reward_1}
    r = & \underbrace{
    \left(1 - e^{ - \lvert \theta _{rad} \rvert}\right)}_{(a)} \\
    & \cdot
    \underbrace{
    \vphantom{\left(1 - e^{ - \lvert \theta _{rad} \rvert}\right)}
    \max(0,~\min(1,~1 - c_0 \cdot (\theta _{deg} - 24) / 3)),}_{(b)}
\end{split}
\end{equation}
where $\theta_{rad}$ and $\theta_{deg}$ are the machine's pitch angle in radians and degrees, respectively, and $c_0$ is a positive coefficient. \autoref{fig:figure_4}B illustrates the machine configuration with different pitch angles.
The term (a) encouraged high pitch values (positive or negative), and (b) discouraged positive pitch values above 24$^\circ$ to avoid balancing on the rear wheels. The coefficient $c_0$ allows the adjustment of the shape of the reward function (\autoref{fig:figure_4}C) and was subject to a curriculum \cite{curriculum} according to \autoref{fig:figure_4}D. The curriculum discouraged touching the ground with the machine's rear underside when switching from driving backward to forward. Because the machine needed to accumulate energy to perform the swing-up maneuver, it had to go on the back wheels (positive pitch angle) with forward velocity. To learn the basic motion quickly with the help of the machine's rear underside, the penalty was only gradually increased as training progressed.

\paragraph{Training}
The swing-up reward (\autoref{fig:figure_4}D) started to increase substantially around iteration 130, which corresponded to the policy learning to lift the front wheels off the ground and stabilize with the help of the rear underside. At around iteration 300, the reward function increased again when the agent learned to swing up onto the front wheels. The average reward then slowly increased as the policy learned how to perform the swing-up maneuver more reliably while avoiding touching the ground with the rear underside. The curriculum on the parameter $c_0$ (\autoref{fig:figure_4}D) increased between iterations 300 and 500 and penalizes large positive pitch angles to discourage the machine's rear underside from touching the ground.

The final performance of the controller was evaluated on 1000 randomized environments. The distribution of the pitch angle at the end of the episode is shown in \autoref{fig:figure_4}E. Every environment performed the swing-up maneuver successfully and ended on the front wheels with a pitch angle below $-30^\circ$ where the stabilize-and-balance controller was activated. The bar at $-52^\circ$ corresponded to the machine landing on the bucket.

\subsubsection*{Stabilize and Balance}
\label{sec:slow_swing_up}
The swing-up controller brought the machine onto its two front wheels into a configuration close to the balancing point. To stabilize the swing-up maneuver and keep the machine balanced on the front wheels without touching the ground with the bucket, we trained another controller.

\begin{figure*}
    \centering
    \includegraphics[width=0.95\linewidth]{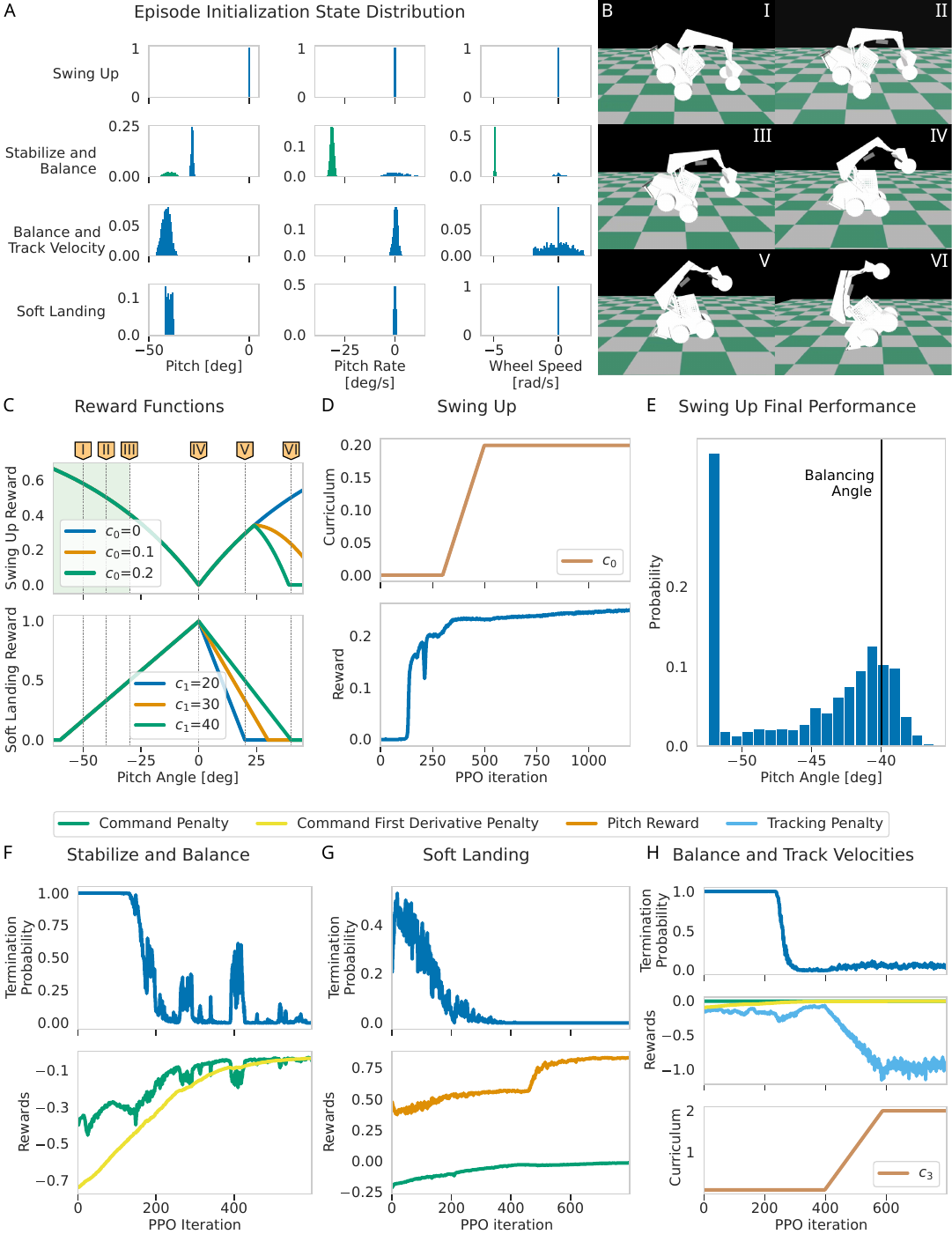}
    \caption{\textbf{RL.}
        \textbf{(A)} Initial state distribution at episode start (sampled on 1000 environments).
        \textbf{(B)} Visualization of different \ac{SSL} pitch angles.
        \textbf{(C)} Rewards as a function of the pitch angle for swing-up and soft-landing controllers for different values of $c_{0,1}$.
        \textbf{(D)} Swing-up learning: Average reward per time step and curriculum progression on $c_0$ to increasingly penalize touching the ground with the rear underside.
        \textbf{(E)} Distribution of the pitch angle at the end of an episode. All 1000 trials achieved a pitch angle smaller than $-30^\circ$, which is required for successful swing-up.
        \textbf{(F)} Stabilize and balance learning: The termination probability drops from 1 to 0, corresponding to the transition from failing all of the time to successful balancing without touching the ground. The rewards for the command and the command derivative increase over the course of training as the controller learns how to perform the task using small and smooth actions. (There is no curriculum for this task.)
        \textbf{(G)} Soft landing learning: The probability of terminating by landing on the bucket or tipping over backward drops to 0. The average reward per time step increases. (There is no curriculum for this task.)
        \textbf{(H)} Balance and track velocities learning: The probability of falling when balancing and tracking velocities drops from 1 to 0. The tracking reward decreases because of the curriculum on $c_3$.
    }
    \label{fig:figure_4}
\end{figure*}

\paragraph{Episode Initialization}
To facilitate exploration, the \ac{SSL} was initialized either near balancing or midswing. Initialization near balancing involved a pitch angle around the balancing angle $(-40^\circ$), a low pitch rate, and low wheel speed (\autoref{fig:figure_4}A), such that the \ac{SSL} started to slowly fall either onto the arm or back onto four wheels if no action was applied. Starting midswing was characterized by a pitch angle around $-30^\circ$, a pitch rate around $-35\si{\degree \per \second}$, and a wheel speed around $5~\si{\radian \per \second}$. Initializing midswing ensured overlapping state distributions between the swing-up and stabilize-and-balance controllers. During deployment, we switched controllers at the midswing pitch value, which lay close to the balancing pitch while maintaining sufficient margin to the $-52^\circ$ pitch where the bucket would touch the ground.

\paragraph{Reward}
The episode terminated with zero reward if the bucket touched the ground or the machine's pitch angle was too far from the balancing point, specifically, larger than $-10^\circ$ (\autoref{fig:figure_4}B, IV to VI).
The reward function was defined as
\begin{equation} \label{eq:swing_up_reward_2}
r =
\underbrace{
    \vphantom{\alpha _{2} \cdot \left( \frac{\partial C}{\partial t} \right)^{2}}
c_{0}}_{(a)}
- \underbrace{
    \vphantom{\alpha _{2} \cdot \left[ \frac{\partial C}{\partial t} \right]_1^2}
c_{1} \cdot a^2}_{(b)}
- \underbrace{
c_{2} \cdot \left[ \frac{\partial a}{\partial t} \right]_1^2}_{(c)
},
\end{equation}
where $c_{0, 1, 2}$ are positive constant reward coefficients, $a$ is the policy action, and $[\partial (\cdot) / \partial t]_1$ is the first-order approximation of the derivative. Here, (a) encouraged the controller to balance continuously, (b) penalized large commands, and (c) incentivized smooth actions.

\paragraph{Training}
\autoref{fig:figure_4}F shows the progress of training. The probability of early termination reached a minimal value after roughly 250 \ac{PPO} iterations. At this point, the controller could successfully slow the swing-up and balance the machine, albeit with high command amplitudes. The training continued until iteration 600, at which point the command penalty and the command derivative penalty converged.

\subsubsection*{Soft Landing}

After balancing, the machine had to be brought safely back onto the ground. If the controller is stopped abruptly (zero command), the landing impact can severely damage the machine. Therefore, we trained a controller that can safely bring the machine from balancing on two wheels to four wheels on the ground, reducing the impact and avoiding rebound.

\paragraph{Episode Initialization}
At the beginning of the episode, the \ac{SSL} was initialized with a pitch angle near the balancing configuration, a low pitch rate, and a small wheel speed (\autoref{fig:figure_4}A). This corresponded to the state when the balancing-and-velocity-tracking controller receives a command of zero desired velocity and was also used as the controller switching configuration during deployment.

\paragraph{Reward}
The episode terminated with 0 reward if the bucket touched the ground or the machine was suspended on its rear underside without any wheel in contact with the ground, which meant a pitch angle above 4$0^\circ$ (\autoref{fig:figure_4}B, VI). The reward function is defined as
\begin{equation} \label{eq:soft_landing_reward}
r = - \underbrace{
\vphantom{\max \left(0, \frac{ \left(\theta _{deg} + c(\theta _{deg})) \right) }{ c(\theta _{deg})} \right)}
 c_{0} \cdot a^{2}}_{(a)}
  +
  \underbrace{
  \max \left(0, \frac{ \left(- \lvert \theta _{deg} \rvert + c(\theta _{deg}) \right) }{c (\theta _{deg})} \right)}_{(b)},
\end{equation}
with
\begin{equation*}
    \begin{aligned}
        c(\theta _{deg}) &= c_{1} &&\text{if} \quad \theta _{deg} \geq 0\\
        c(\theta _{deg}) &= c_{2} &&\text{if} \quad \theta _{deg} < 0,
    \end{aligned}
\end{equation*}
where $\theta_{deg}$ is the machine's pitch in degrees, $c_{0, 1, 2}$ are positive reward coefficients, and $a$ is the command. Here, (a) encouraged small actions and (b) incentivized a soft landing without rebounding.

\autoref{fig:figure_4}C shows element (b) of the soft-landing reward and highlights the influence of the coefficient $c_1$. Note that $c_1$ was constant during training and not subject to a curriculum. The reward increased linearly with the pitch angle until it reached $0^\circ$. Positive pitch angles were then penalized to discourage rebounding with the front wheels.

\paragraph{Training}
\autoref{fig:figure_4}G shows the progress of training. After 300 iterations, the agent learned to avoid terminations (such as landing on the bucket or tipping over backward) and consistently landed on the rear wheels.
Until iteration 460, the agent tried to bring the rear wheels to the ground as quickly as possible by accelerating forward. This led to a strong impact and rebound of the front wheels. The pitch reward increased slowly during this stage. After iteration 460, the agent found that the rebound could be avoided by accelerating backward when the rear wheels were close to the ground, which led to a notable increase in pitch reward. Throughout training, the command penalty decreased as the agent learned to apply smaller command magnitudes.

\subsubsection*{Balance and Track Velocities}

When the \ac{SSL} was stabilized after swing-up in a balancing configuration, the goal was to track desired velocity commands to move the machine. We trained a controller to track linear velocities, specifically the desired average wheel speed.

\paragraph{Episode Initialization}
The training episode was initialized with a randomized pitch angle around the balancing angle, a low pitch rate, and nonzero wheel speed (\autoref{fig:figure_4}A). This corresponds to the state after executing the stabilize-and-balance controller and was also used as a criterion to activate this controller. Although the wheel speed was low after swing-up (for example, $\pm0.3$~\si{\radian\per\second}), the initial wheel speed was sampled within $\pm2$~\si{\radian\per\second}, to support exploration and increase robustness. The target wheel speed was observed and uniformly sampled within $\pm4$~\si{\radian \per \second}. Furthermore, the target was updated during the episode to expose the agent to arbitrary combinations of measured and commanded speed. The update time was sampled between 4.2 and 5~\si{\second}.

\paragraph{Reward}
The episode terminated with zero reward if the bucket touched the ground or when the chassis pitch angle was above $-10^\circ$. The reward function was defined as
\begin{equation}
\label{eq:vel_tracking_reward}
\begin{split}
r =& \underbrace{ \vphantom{\alpha _{2} \cdot \left( \frac{\partial C}{\partial t} \right)^{2}} c_{0}}_{(a)}
- \underbrace{ \vphantom{\alpha _{2} \cdot \left( \frac{\partial C}{\partial t} \right)^{2}} c_{1} \cdot a^2}_{(b)}
- \underbrace{ c_{2} \cdot \left[ \frac{\partial a}{\partial t} \right]_1^2}_{(c)} \\
&- \underbrace{ \vphantom{\alpha _{2} \cdot \left( \frac{\partial a}{\partial t} \right)^{2} } c_{3} \cdot \left( u - u^* \right) ^2}_{(d)},
\end{split}
\end{equation}
where $c_{0, 1, 2, 3}$ are positive coefficients, $a$ is the command, and $u$ and $u^*$ are the measured and desired wheel speeds, respectively. Here, (a) encouraged the controller to balance continuously, (b) encouraged small actions, (c) encouraged smooth actions, and (d) rewarded tracking the desired average wheel speed.

\paragraph{Training}
\autoref{fig:figure_4}H summarizes the progress of training. The agent learned to balance the \ac{SSL} between iterations 240 and 300, indicated by the dropping termination probability (\autoref{fig:figure_4}H, Termination Probability).
\autoref{fig:figure_4}H (Rewards), shows the values of the reward components. The rewards on the command and the command derivative increased as training progressed, indicating that the controller learned to use smaller and smoother actions. Note that the value of these penalties was much smaller than for the stabilizing and balancing controller. There, the goal was to balance in place. Here, in contrast, the policy needed to be able to apply more aggressive commands when the velocity target changed, such as when going from backward motion to forward motion. Relaxing the command and command-derivative constraints allowed the policy to track the provided velocities. To facilitate the learning process, we used a curriculum on the tracking reward by increasing the reward coefficient $c_3$ (\autoref{fig:figure_4}H, Curriculum). When training began, the tracking penalty was small, and the agent could focus on learning to balance. Once the agent learned to balance, we increased the penalty on the wheel speed between iterations 400 and 590 to encourage more accurate velocity tracking. Although the tracking reward improved only slightly between iterations 600 and 800, the final tracking performance improved substantially.

Akin to the other controllers described here, this controller could be deployed directly on the physical hardware without filtering. However, its behavior could be improved by filtering the policy actions ($a_t$) before applying them to the machine with a simple low-pass filter
\begin{equation} \label{eq:filter}
a_{\textup{filt}, t} = \beta \cdot a_{\textup{filt}, t-1} + \left(1-\beta\right) \cdot a_{t},
\end{equation}
where $a_{\textup{filt}}$ is the filtered action and $\beta = 0.8$ is an empirically tuned coefficient.

\subsection*{Statistical Analysis}

Each controller was evaluated in simulation using 300 episodes executed in parallel. The reported reward is the arithmetic mean of the cumulative discounted rewards across all episodes. The termination probability is the fraction of rollouts that failed to achieve the task objective. Initial and final-state distributions were estimated using 1000 parallel simulation environments.

On real hardware, each \ac{RL} controller was compared with its simulated counterpart and with the relevant baseline controllers by overlaying sensor time series. The baselines were the \ac{MPC} controller and, for the soft-landing controller, a zero-command baseline.
The actuator model was evaluated using prediction errors on training and validation datasets obtained on the real machine. To characterize the magnitude and variability of these errors, we computed the mean absolute error, standard deviation, and 95th percentile of the error distribution. In addition, the temporal lag was evaluated to assess the model’s ability to capture actuation delays.

\section*{Acknowledgments}
We thank A.~Marsh for essential technical support, including machine maintenance and hardware debugging; D.~Fernandez for assisting with the experiments and always providing a helping hand; and R.~Welten for providing and managing the testing infrastructure. We also thank T.~O’Brien and D.~Zandman for professional drone operation and filming. Last, we thank J.~Ostrowski for comprehensive project support.

\section*{Funding}
This work was supported by the Swiss National Science Foundation grant.%

\section*{Competing Interests}
The authors declare that they have no competing interests.

\section*{Data, Code, and Materials Availability}
Datasets and code to generate all figures and supplementary figures are made publicly available at \href{https://doi.org/10.5061/dryad.d7wm37qhh}{10.5061/dryad.d7wm37qhh}.

\section*{Disclaimer}
This work was performed for research purposes only. Nothing in this paper constitutes an endorsement by Blue River Technologies (BRT) or John Deere to attempt these operations of heavy machinery, either autonomously or by human operators, outside of this specific experimental framework.

\section*{Supplementary Material}
\noindent
\begin{minipage}[t]{0.29\columnwidth}
\ref{supp:MPC}
\end{minipage}
\begin{minipage}[t]{0.69\columnwidth}
\ac{SSL} Model Predictive Control Baseline
\end{minipage}
\newline

\noindent
\begin{minipage}[t]{0.29\columnwidth}
\ref{sup_sec:smb}
\end{minipage}
\begin{minipage}[t]{0.69\columnwidth}
Case Study: Super Mega Bot (SMB)
\end{minipage}
\newline

\noindent
\begin{minipage}[t]{0.29\columnwidth}
\ref{tab:ssl_actuator_model_data_collection}
\end{minipage}
\begin{minipage}[t]{0.69\columnwidth}
\ac{SSL} Actuator Model Dataset
\end{minipage}
\newline

\noindent
\begin{minipage}[t]{0.29\columnwidth}
\ref{tab:ssl_act_model_eval}
\end{minipage}
\begin{minipage}[t]{0.69\columnwidth}
\ac{SSL} Actuator Model Evaluation
\end{minipage}
\newline

\noindent
\begin{minipage}[t]{0.29\columnwidth}
\ref{tab:ssl_observed_variables}
\end{minipage}
\begin{minipage}[t]{0.69\columnwidth}
\ac{SSL} Policy Observations
\end{minipage}
\newline

\noindent
\begin{minipage}[t]{0.29\columnwidth}
\ref{tab:ssl_dr}
\end{minipage}
\begin{minipage}[t]{0.69\columnwidth}
\ac{SSL} Rigid-Body Model Randomization
\end{minipage}
\newline

\noindent
\begin{minipage}[t]{0.29\columnwidth}
\ref{tab:ssl_external_disturbance}
\end{minipage}
\begin{minipage}[t]{0.69\columnwidth}
\ac{SSL} Disturbance Forces
\end{minipage}
\newline

\noindent
\begin{minipage}[t]{0.29\columnwidth}
\ref{tab:SSL_training}
\end{minipage}
\begin{minipage}[t]{0.69\columnwidth}
\ac{SSL} Policy Training Parameters
\end{minipage}
\newline

\noindent
\begin{minipage}[t]{0.29\columnwidth}
\ref{tab:ssl_reward_function_paramters}
\end{minipage}
\begin{minipage}[t]{0.69\columnwidth}
\ac{SSL} Reward Functions Parameters
\end{minipage}
\newline

\noindent
\begin{minipage}[t]{0.29\columnwidth}
\ref{tab:mpc_cost_function}
\end{minipage}
\begin{minipage}[t]{0.69\columnwidth}
\ac{SSL} \ac{MPC} Parameters
\end{minipage}
\newline

\noindent
\begin{minipage}[t]{0.29\columnwidth}
\ref{tab:smb_act_model_eval}
\end{minipage}
\begin{minipage}[t]{0.69\columnwidth}
SMB Actuator Model Evaluation
\end{minipage}
\newline

\noindent
\begin{minipage}[t]{0.29\columnwidth}
\ref{tab:smb_observed_variables}
\end{minipage}
\begin{minipage}[t]{0.69\columnwidth}
SMB Policy Observations
\end{minipage}
\newline

\noindent
\begin{minipage}[t]{0.29\columnwidth}
\ref{tab:smb_param_rand}
\end{minipage}
\begin{minipage}[t]{0.69\columnwidth}
SMB Rigid-Body Model Randomization
\end{minipage}
\newline

\noindent
\begin{minipage}[t]{0.29\columnwidth}
\ref{tab:smb_act_noise}
\end{minipage}
\begin{minipage}[t]{0.69\columnwidth}
SMB Actuator Perturbation
\end{minipage}
\newline

\noindent
\begin{minipage}[t]{0.29\columnwidth}
\ref{tab:smb_reward}
\end{minipage}
\begin{minipage}[t]{0.69\columnwidth}
SMB Reward Function Coefficients
\end{minipage}
\newline

\noindent
\begin{minipage}[t]{0.29\columnwidth}
\ref{tab:smb_training}
\end{minipage}
\begin{minipage}[t]{0.69\columnwidth}
SMB Policy Training Parameters
\end{minipage}

\bibliography{scibib}

\bibliographystyle{sciencemag-arxiv}

\clearpage

\section*{Supplementary Materials}
\renewcommand\thesection{Section S}
\renewcommand{\subsectionautorefname}{}

\subsection{SSL Model Predictive Control Baseline}
\label{supp:MPC}
We used \ac{MPC} as a reference baseline for the presented \ac{RL} approach. \Ac{MPC} is an advanced control methodology that solves an optimization problem using a mathematical model to predict the system's behavior. \Ac{MPC} has been applied in many domains, including robotics, and often outperforms classic feedback controllers \cite{rawlings2017model, mattila_survey, kalmari_forestry, Prado2021_skid_steer}.

\subsubsection*{Model}

\ifImageAtTheEnd \else
\begin{figure}
\centering
  \includegraphics[width=\linewidth]{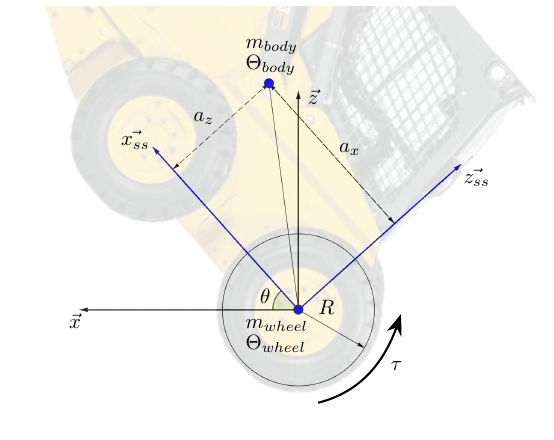}
  \caption{\textbf{Mathematical description of the \ac{SSL} used for the \ac{MPC}.} Each front wheel is modeled as a cylindrical body with radius $R$, mass $m_{wheel}$, and inertia $\Theta_{wheel}$. The rest of the machine is lumped into a body of mass $m_{body}$ and inertia $\Theta_{body}$. }
  \label{fig:MPC_description}
\end{figure}
\fi

The \ac{SSL} was modeled as a planar inverted pendulum, where the front wheels are cylinders, and the rest of the machine is lumped into a mass with inertia around the $y$-axis (\autoref{fig:MPC_description}). The model parameters were determined using truck scales and geometric approximations, similar to the \ac{RL} approach. We omitted modeling of tire dynamics in both the \ac{RL} simulation and the \ac{MPC} model, as these dynamics are primarily relevant for skidding maneuvers like path following on all four wheels~\cite{Prado2021_skid_steer}.

The state $\V x \in \mathbb{R}^4$ and input $\V u \in \mathbb{R}^1 $ are defined as
\begin{equation}
    \V x = [x, v, \theta, \dot{\theta}]^T
\end{equation}
\begin{equation}
    \V u = [u]^T,
\end{equation}
where $x$ and $v$ are the wheel center position and velocity, $\theta$ and $\dot{\theta}$ the machine's pitch angle and angular velocity, and $u$ the desired wheel speed.

Although data from the real machine was used to learn an actuator model for the \ac{RL} approach, the actuators were modeled as a first-order system for the \ac{MPC}. Since the \ac{SSL} is velocity controlled (\autoref{fig:figure_3}B), the actuation torque $\tau$ was computed in proportion to the error between measured and desired wheel speed as
\begin{equation}
    \tau = \frac{u - (\dot{\theta} + \frac{v}{R})}{T},
\end{equation}
where $u$ is the desired wheel speed, $\dot{\theta}$ the machine's pitch rate, $R$ the wheel radius, and $T$ the time constant of the actuators. The time constant is identified via a system identification procedure.

The following set of equations describes the dynamics of the \ac{SSL}:
\begin{equation}
    \begin{aligned}
        \dot{x} &= v \\
        \dot{v} &= \frac{(-g \cdot A_1 + \frac{A_4}{A_3} \cdot (A_1 \cdot \dot{\theta} ^ 2 - \frac{\tau}{R}) - \tau)}{A_3 - \frac{A_4 \cdot A_3}{A_2}} \\
        \dot{\theta} &= \dot{\theta} \\
        \ddot{\theta} &= \frac{A_1 \cdot \dot{\theta}^2 + A_2 \cdot \dot{v} - \frac{\tau}{R}}{A_3},
    \end{aligned}
\end{equation}
with
\begin{equation}
    \begin{aligned}
        A_1 &= m_{body} \cdot (a_z \cdot \sin{\theta} - a_x \cdot \cos{\theta}) \\
        A_2 &= m_{body} + 2 \cdot m_{wheel} + 2 \cdot \Theta_{wheel}R^{-2} \\
        A_3 &= m_{body} \cdot (a_x \cdot \sin{\theta} + a_z \cdot \cos{\theta}) \\
        A_4 &= \Theta_{body} + m_{body} \cdot (a_x^2 + a_z^2). \\
    \end{aligned}
\end{equation}

\subsubsection*{Cost Function}
The cost function is
\begin{equation}
    L =\V x^T \V Q \V x + \V u^T \V R \V u + \dot{\V u}^T \V R _d \dot{\V u},
\end{equation}
where $\V Q = \textup{diag}(q_x, q_v, q_{\theta}, q_{\dot{\theta}})$. $\V R$ and $\V R _d$ are tuning parameters.

The tracking is performed using the following change of variables:
\begin{equation}
    \begin{aligned}
        \V x &\coloneqq \V x_{meas} - \V x_{des} \\
        \V u  &\coloneqq  \V u_{meas} - \V u_{des}.
    \end{aligned}
\end{equation}
For static balancing, $\V x_{des} = [0, 0, \theta_{bal}, 0]^T$. For balancing while tracking a target velocity, $\V x_{des} = [0, v_{des}, \theta_{bal}, 0]^T$. Thus, $\theta_{bal}$ is the machine's pitch angle when balancing on the front wheels and its value was refined on the fly to achieve better performance. Note that for velocity tracking, no specific targets were specified for $x$ and $\dot{\theta}$ because the corresponding costs are $0$. The cost terms and time constant are specified in \autoref{tab:mpc_cost_function}.

The optimization problem was solved using CasADi \cite{casadi} with a horizon of $40$~steps and a step size of $0.02~\si{\second}$. For the deployment on the real machine, the desired wheel speed $u$ was converted to a machine command by interpolating values from a lookup table consisting of $10$~speed-command value pairs that have been pre-recorded.

With this simple \ac{MPC} baseline, swing-up or soft landing could not be achieved due to the contact changes required by these maneuvers. For testing, the machine was brought on 2 wheels and back on 4 wheels manually by a human operator.
Prior work on legged systems has adapted MPC to complex motions with contact switches, but had to resort to fixed gait patterns~\cite{grandia2023perceptive} or dedicated sampling-based gait planners~\cite{jelavic2021combined,jelavic2023lstp}.

\subsection{Case Study: Super Mega Bot (SMB)}
\label{sup_sec:smb}

\ifImageAtTheEnd \else
\begin{figure*}
   \centering
   \includegraphics[width=0.85\linewidth]{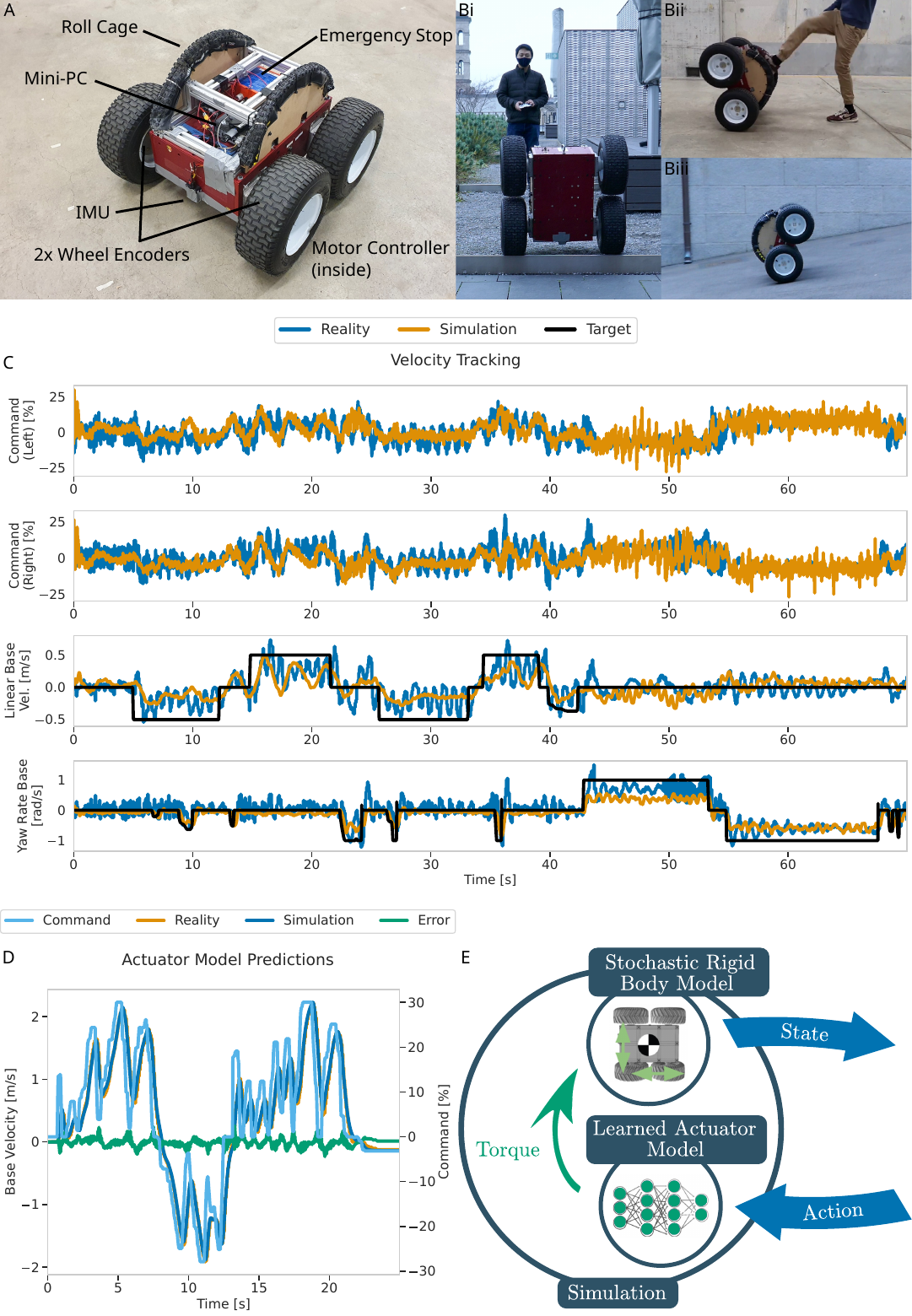}
   \caption{\textbf{\Acf{SMB}.}
   \textbf{(A)} The \ac{SMB} is a skid-steer vehicle for heavy-duty use in challenging conditions. It has been retrofitted with a 6-\acs{DoF} \ac{IMU}, a PC to run the policy, read the \ac{IMU} signals and communicate with the motor controller, an emergency stop for safety, a roll cage for protection of the components, and wheel encoders on the front wheels.
   \textbf{(Bi)} The operator is navigating the \ac{SMB} accurately along the path by providing high-level commands consisting of linear and angular velocity. The policy is robust to uneven ground and driving down small steps. \textbf{(Bii)} The policy can reject external pushes, which can cause the \ac{SMB} to travel a couple of meters. \textbf{(Biii)} the policy is robust to non-flat ground and can successfully balance and follow tracking commands.
   \textbf{(C)} Control command and velocity tracking performance in simulation and on the real system while balancing, given the same command sequence.
   \textbf{(D)} The hybrid simulation of the \ac{SMB} consists of an actuator model trained to predict torques based on real-world data, and the rigid-body model whose properties are estimated and randomized around the measured values during training.
   \textbf{(E)} Actuator model performance evaluation on a validation dataset in simulation compared to measurements on the real machine.
   }
   \label{fig:smb_full}
\end{figure*}
\fi

\subsubsection*{Results}
To demonstrate the generality of the presented methodology, we applied it to the Super Mega Bot (SMB), an electric, heavy-duty four-wheel drive (4WD) platform with a similar topology to the skid-steer loader. The \ac{SMB} weighs 100~\si{\kg} and has a towing capacity of 1.25~tonnes~\cite{inspectorbots} (\autoref{fig:smb_full}A).

As with the \ac{SSL}, we trained the \ac{SMB} in simulation to balance on two wheels while tracking user-defined linear and angular base velocities. The learned policy was then deployed on the real machine without any further modifications.
When no velocity was commanded, the policy kept the \ac{SMB} balanced in place. Upon receiving a target linear or angular velocity, the robot followed the given commands quickly, without introducing unwanted motions, such as linear velocities when tracking angular velocities, or vice versa. This decoupling allowed users to maneuver the \ac{SMB} around easily.
The learned policy also showed robustness to a variety of disturbances. It could reject external pushes from random directions, some of which caused the robot to travel several meters before regaining its balance. Similar to the \ac{SSL}, the policy exhibited generalization capabilities for which it has not been explicitly trained. For example, it could traverse uneven terrain and rough surfaces such as gravel, and even recover from driving over steps (\autoref{fig:smb_full}B). Although the \ac{SMB} is smaller than the \ac{SSL}, the control task is more challenging due to the lower inertia of the vehicle, which demands faster and more precise control inputs. The experimental results are showcased in Movie S6.

To enable autonomous operation, minor hardware modifications were made to the original platform (\autoref{fig:smb_full}A). The two front wheels were equipped with encoders, and a 6-\acs{DoF} \ac{IMU} was attached to the robot's base. A mini PC (Intel Core i7 CPU, 8~GB RAM, no GPU) was used to interface the motor controller and \ac{IMU} via a \ac{CAN} bus and run the control policy. To protect the hardware and ensure safe operation, the robot was equipped with a roll cage and a radio emergency stop.

To test the policy, the \ac{SMB} was manually lifted into an upright configuration from which the controller is started. \autoref{fig:smb_full}C compares the policy's control signals, as well as the commanded and measured linear and angular velocities, both on the real system and in simulation.
In the absence of external commands, the policy kept the robot balanced without noticeable drift in either linear or angular velocity. To stabilize the machine on two wheels, the policy performed small back-and-forth motions. Compared to the simulation, they were of a slightly larger magnitude and lower frequency. We attribute this to the residual sim-to-real gap.

When commanded to follow a specific linear and angular velocity, the robot responded in the intended direction. However, when comparing the target commands with the estimated velocities, it becomes apparent that the policy did not track the target commands accurately and instead moved consistently slower than it was supposed to. Notably, this discrepancy was also present in simulation, suggesting that more accurate tracking is possible by further shaping the reward.

\subsubsection*{Actuator Model}

For electric drives, the control signal applies to the motor torque. Therefore, instead of predicting wheel speed, the actuator model was trained to map a history of control signals and wheel velocities up to time step $t$ to the torque at time step $t+0.01~\si{\second}$ (\autoref{fig:smb_full}D). The history was needed since the actuator is not perfectly stateless due to hysteresis effects and delays. However, the motor torque at the wheel cannot be measured directly. We approximated the wheel torque based on the assumption that on flat ground and in the absence of wheel slippage, the total torque $\tau_\textrm{tot}$ generated by the four wheels was given by the acceleration in driving direction $a_x$, the mass of the robot $m_\textrm{robot}$, and the wheel radius $r_\textrm{wheel}$, as summarized in \autoref{eq:total_torque}. Further, to derive the torque per wheel $\tau_{wheel}$, we assumed that the contribution of each wheel to the acceleration is constant (\autoref{eq:wheel_torque}).
\begin{subequations}
\begin{align}
    \tau_\textrm{tot} &\approx m_\textrm{robot} \;a_x \;r_\textrm{wheel} \label{eq:total_torque} \\
    \tau_\textrm{wheel} &\approx \frac{\tau_\textrm{tot}}{4}  \label{eq:wheel_torque}
\end{align}
\end{subequations}

Data for training and validating the actuator model was collected with a remote control by driving back and forth at various speeds and accelerations for 5.5~\si{\min} on flat ground with up to 30\% of the maximum command. This avoided wheel slippage, reduced the required amount of training data and was sufficient for balancing. An excerpt of the data collection is shown in Movie S6. The history length chosen for the actuator model has been set to 30 time steps, with each step lasting 0.01~\si{\second}, which is above the maximum observed delay. The neural network consists of three hidden layers with 32 nodes, SoftSign activation, and a linear output layer. It was trained using the MSE loss and the Adam Optimizer~\cite{adam_opt} until the validation loss stopped improving.
\ref{tab:smb_act_model_eval} summarizes the quantitative evaluation of the learned actuator model.

\subsubsection*{Rigid-Body Model}

Since no robot model was available from the manufacturer, the shape of the robot was approximated as a square main body with four cylindrical wheels with uniform density. Note that the roll cage is not part of the collision geometry, as it served purely as a safety device and must not be exploited by the agent during training. It is, however, included in the center-of-mass estimation.
The dimensions, positions, and masses were measured with a tape measure and a weighing scale. The center of mass of the wheels was assumed to be in the middle of the cylinder. For the base, the three dimensions of the center of mass were estimated with two scales via equilibrium of moments~\cite{omnicalculator}. The inclination for estimating the height of the center of mass was obtained from the \ac{IMU}.

\autoref{fig:smb_full}E shows an excerpt of the robot's base velocity in the real world as well as in simulation, including the learned actuator dynamics and the rigid body model for the same command sequence for a dataset not used for training the actuator model. The simulation aligned well with reality and was also able to capture delays accurately.

\subsubsection*{Observations and Actions}
The agent observed
the measured acceleration in all three axes of the \ac{IMU},
the measured angular velocity around all three axes of the \ac{IMU},
the pitch angle, calculated by the internal extended Kalman filter of the \ac{IMU},
previous action,
and the target commands for linear and angular velocities.
To allow the policy to infer the robot's dynamics implicitly from noisy measurements and handle delays, a history of the above measurements was provided as input to the policy. The history length was chosen to match the history of the actuator model, specifically, 15 steps at a frequency of 50~\si{\hertz}. Since policies trained and deployed at 100~\si{\hertz} performed similarly, we selected 50~\si{\hertz} to reduce training times.
The observations from the \ac{IMU} and the wheel encoders were modeled according to the specifications in the manufacturers' data sheets. The models have the same resolution, saturation, and noise level as their real counterparts. The noise has been modeled as a normal distribution around the true value. The numerical values and sensor details are summarized in \ref{tab:smb_observed_variables}.

The action consists of two signals sent directly to the motor controller for either side of the robot. Actions and observations were not scaled as their magnitudes and distributions are similar.

\subsubsection*{Episode Initialization}
At the beginning of every episode, the \ac{SMB} was initialized with a pitch angle sampled from a uniform distribution within $\pm20^\circ$ around the experimentally determined balancing pitch angle.

\subsubsection*{Domain Randomization and External Disturbances}
Domain randomization was applied to the mass, center of mass, and moment of inertia. The values were sampled from a uniform distribution according to the values listed in \autoref{tab:smb_param_rand}.

Furthermore, to prevent the policy from overfitting to the approximate actuator model trained on real-world data, a random torque was added to each actuator, sampled from a uniform distribution. This torque was sampled anew with a certain probability for each time step and persisted for a few time steps or up to several seconds, making the policy robust against both high- and low-frequency noise. Effectively, this corresponds to external disturbances. The numerical values are listed in \autoref{tab:smb_act_noise}.

\subsubsection*{Reward}

The reward function $r_t$, used in each time step $t$, is given in \autoref{eq:smb_rew}:
\begin{equation}
\label{eq:smb_rew}
\begin{split}
    r_t = &\underbrace{c_1}_{(a)}
    - \underbrace{c_2\;(\vert a_t^l \vert + \lvert a_t^r \rvert) }_{(b)}
    - \underbrace{c_3 \; \lvert v_t^* - v_t \rvert}_{(c)} \\
    &- \underbrace{c_4\; \lvert \omega_t^* - \omega_t \vert.}_{(d)}
\end{split}
\end{equation}
Here $c_{1-4}$ are positive coefficients, $a_l$ and $a_r$ are the policy actions for the left- and right-side wheels, respectively, and $v$ and $\omega$ are the scalar linear and angular base velocities, respectively. The reward components (a) discouraged termination via a constant per-step reward, (b) encouraged the agent to use small commands for the left- and right-side wheels, (c) penalized deviations from the desired scalar linear base velocity, and (d) penalized deviations from the desired scalar base yaw velocity.

Since a high negative reward could outweigh the positive reward and result in a negative overall reward, the trained policy might learn to end every episode as soon as possible if it cannot find a way to achieve positive rewards. To prevent this, a curriculum \cite{curriculum} was implemented that kept the costs on tracking the target commands negligibly small at first and increased them with every training iteration up to a maximum. Specifically, $c_3$ and $c_4$ increased with every iteration until $c_3=c_{3\textrm{,max}}$ and $c_4=c_{4\textrm{,max}}$. The effect was that the policy first learned to balance and then gradually learned to follow the target commands. The specific parameters of the reward function are listed in \autoref{tab:smb_reward}. The policy training parameters are specified in \autoref{tab:smb_training}.

\setcounter{subsection}{0}
\refstepcounter{section}
\renewcommand{\thesection}{Table S}

\subsection{SSL Actuator Model Dataset}
\label{tab:ssl_actuator_model_data_collection}

The dataset contains 5 types of signals. The sinusoids have amplitudes between $-30~\si{\percent}$ and $30~\si{\percent}$ with increments of $10~\si{\percent}$ and the frequency ranges between $0.2~\si{\hertz}$ and $1.67~\si{\hertz}$. The ramps amplitude range between $-70~\si{\percent}$ and $70~\si{\percent}$ with increments of $10~\si{\percent}$ and a rise time between $0.5~\si{\second}$ and $2.85~\si{\second}$. The steps have amplitudes between $-100~\si{\percent}$ and $100~\si{\percent}$ with increments of $5~\si{\percent}$. The dataset is standardized by subtracting the means and dividing by the standard deviations.

\begin{table}[H]
\small\sf\centering
 \begin{adjustbox}{max width=\columnwidth}
\begin{tabular}{@{}lllllll@{}}
\hline
& \textbf{Sinusoids} & \textbf{Ramps} & \textbf{Steps} & \textbf{Balancing} & \textbf{Zero} & \textbf{Total} \\
\hline
Duration~$[\si{\minute}]$ & $20$ & $25$ & $27$ & $20$ & $15$ & $107$ \\
\hline
\end{tabular}
\end{adjustbox}
\end{table}

\subsection{SSL Actuator Model Evaluation}
\label{tab:ssl_act_model_eval}

The measurement rate is 50~\si{\hertz}, and the actuator model predicts the velocity at the same rate. The observation histories consist of 30 datapoints, sampled sparsely with stride of 3 and cover the past 1.8~\si{\second}. The validation data consists of 2 datasets: one is collected while running a controller that balances the machine (Validation 1) and the other is collected manually, driving the machine back and forth on 4 wheels (Validation 2). Before training the model, the datasets were standardized by subtracting the mean and dividing by the standard deviation. In the following table, we report the Mean Average Error (Mae), Standard Deviation (Std), 95th percentile (p95), and Delay estimation (Lag) computed using normalized cross-correlation between the measured base velocity in simulation and reality. The lag between the command and the measured wheel speed is 0.24~\si{\second}.

\begin{table}[H]
\small\sf\centering
\begin{adjustbox}{max width=\columnwidth}
\begin{tabular}{@{}llllll@{}}
\hline
\textbf{Dataset} & \textbf{Duration $[\si{\second}]$} & \textbf{Mae $[\si{\radian\per\second}]$} & \textbf{Std $[\si{\radian\per\second}]$} & \textbf{p95 $[\si{\radian\per\second}]$} & \textbf{Lag $[\si{\second}]$} \\
\hline
Training & 6420 & 0.0401 & 0.0797 & 0.1490 & 0.0 \\
Validation 1 & 184 & 0.0842 & 0.1397 & 0.2350 & 0.0 \\
Validation 2 & 478 & 0.0846 & 0.1225 & 0.2464 & -0.02 \\
\hline
\end{tabular}
\end{adjustbox}
\end{table}

\subsection{SSL Policy Observations}
\label{tab:ssl_observed_variables}%
Policy observations for different tasks. The history is sparsely sampled, for example, a history of 30 steps and stride 3 covers 1.8~\si{\second} of past measurements given the control rate of 50~\si{\hertz}. The noise is sampled from a Gaussian distribution with mean $\mu = 0$ and standard deviation $\sigma$. The pitch and pitch rate are measured with an Aceinna MTLT305E/D \ac{IMU}~\cite{aceinna}. Wheel speed is obtained from the machine’s onboard sensing system via the \ac{CAN} bus.
\begin{table}[H]
\small\sf\centering
\begin{adjustbox}{max width=\columnwidth}
\begin{tabular}{@{}llllll@{}}
\hline
\textbf{Observation} & & \textbf{Swing-Up} & \textbf{Stabilize and} & \textbf{Balance and} & \textbf{Soft Landing} \\
& & & \textbf{Balance} &\textbf{Track Velocities} & \\
\hline
& History & 30 & 30 & 0 & 0 \\
\textbf{Pitch} $[\si{\radian}]$ & Stride  & 3 & 3 & 0 & 0 \\
& Noise $\sigma$ & 0.01 & 0.01 & 0.01 & 0.01 \\
\hline
& History & 30 & 30 & 0 & 0 \\
\textbf{Pitch rate} $[\si{\radian\per\second}]$ & Stride  & 3 & 3 & 0 & 0 \\
& Noise $\sigma$  & 0.01 & 0.01 & 0.01 & 0.01 \\
\hline
& History & 30 & 30 & 0 & 0 \\
\textbf{Wheel speed} $[\si{\radian\per\second}]$ & Stride  & 3 & 3 & 0 & 0 \\
& Noise $\sigma$ & 0.01 & 0.01 & 0.01 & 0.01 \\
\hline
& History & 30 & 30 & 30 & 30 \\
\textbf{Prev. action} $[\si{\percent}]$ & Stride  & 3 & 3 & 3 & 3 \\
& Noise $\sigma$ & 0.01 & 0.0 & 0.0 & 0.0 \\
\hline
\textbf{Target wheel} & History & -- & -- & 0 & -- \\
\textbf{speed} $[\si{\radian\per\second}]$ & Stride & -- & -- & 0 & -- \\
& Noise $\sigma$ & -- & -- & 0.0 & -- \\
\hline
\end{tabular}
\end{adjustbox}
\end{table}

\subsection{SSL Rigid-Body Model Randomization}
\label{tab:ssl_dr}

The rigid-body randomization parameters are sampled
from a uniform distribution within ± the values listed in this
table. The arm angle is defined in \autoref{fig:hardware_description} and the center of mass
location ($a_x, a_z$) in \autoref{fig:MPC_description}.

\begin{table}[H]
\begin{adjustbox}{max width=\columnwidth}
\begin{tabular}{@{}lllll@{}}
\hline
& \textbf{Swing-Up} & \textbf{Stabilize and} & \textbf{Balance and} & \textbf{Soft Landing} \\
& & \textbf{Balance} & \textbf{Track Velocities} & \\
\hline
$a_x~[\si{\m}]$ & $0.002$ & $0.002$ & $0.002$ & $0.002$ \\
$a_z~[\si{\m}]$ & $0.02$ & $0.02$ & $0.02$ & $0.02$ \\
Body inertia $\Theta_{body}~[\si{\percent}]$ & $10$ & $10$ & $10$ & $10$ \\
Arm angle $[\si{\deg}]$ & $1$ & $1$ & $1$ & $1$ \\
\hline
\end{tabular}
\end{adjustbox}
\end{table}

\subsection{SSL Disturbance Forces}
\label{tab:ssl_external_disturbance}

The disturbance forces are sampled from a uniform distribution within $\pm$ the maximum force and resampled at the given resampling time.

\begin{table}[H]
\small\sf\centering
\begin{adjustbox}{max width=\columnwidth}
\begin{tabular}{@{}lllll@{}}
    \hline
    & \textbf{Swing-Up} & \textbf{Stabilize and} & \textbf{Balance and} & \textbf{Soft Landing} \\
    & & \textbf{Balance} & \textbf{Track Velocities} & \\
    \hline
    Maximum force $[\si{\newton}]$ & $1000$ & $1000$ & $1000$ & $1000$ \\
    Resampling time $[\si{\second}]$ & $3$ & $3$ & $3$ & $3$ \\
    \hline
 \end{tabular}
 \end{adjustbox}
\end{table}

\subsection{SSL Policy Training Parameters}
\label{tab:SSL_training}

\begin{table}[H]
\small\sf\centering
\begin{adjustbox}{max width=\columnwidth}
\begin{tabular}{@{}lllll@{}}
\hline
\textbf{Parameter} & \textbf{Swing-Up} & \textbf{Stabilize and} & \textbf{Balance and} & \textbf{Soft Landing} \\
& & \textbf{Balance} & \textbf{Track Velocities} & \\
\hline
Policy net. & 128x128 & 128x128 & 128x128 & 128x128 \\
Critic net. & 128x128 & 128x128 & 128x128 & 128x128 \\
Nonlinearity & tanh & tanh & tanh & tanh \\
\ac{PPO} iterations & 1200 & 800 & 800 & 800 \\
Batch size & 90k & 120k & 120k & 75k\\
Episode length [\si{\second}] & 6 & 8 & 8 & 5 \\
Control rate [\si{\hertz}] & 50 & 50 & 50 & 50 \\
Discount factor & 0.99 & 0.99 & 0.995 & 0.99 \\
Entropy coeff. & 0.0 & 0.0 & 0.0 & 0.0 \\
Learning rate & 3e-4 & 3e-4 & 3e-4 & 3e-4 \\
Epochs & 4 & 4 & 4 & 4 \\
Mini-batches & 4 & 4 & 4 & 4 \\
\hline
\end{tabular}
\end{adjustbox}
\end{table}

\subsection{SSL Reward Functions Parameters}
\label{tab:ssl_reward_function_paramters}

A curriculum is denoted as  a\textrightarrow b.

\begin{table}[H]
\small\sf\centering
\begin{adjustbox}{max width=\columnwidth}
\begin{tabular}{@{}lllll@{}}
\hline
 & \textbf{Swing-Up} & \textbf{Stabilize and} & \textbf{Balance and} & \textbf{Soft Landing} \\
 && \textbf{Balance} & \textbf{Track Velocities} & \\
\hline
$c_0$  & 0\textrightarrow 0.2 &  4 & 3 & 0.2 \\
$c_1$ & -& 0.5 & 0.01 & 30 \\
$c_2$ & -& 2 & 0.2 & 60 \\
$c_3$ & -& - & 0\textrightarrow 2.0 & - \\
\hline
\end{tabular}
\end{adjustbox}
\end{table}

\subsection{MPC Parameters}
\label{tab:mpc_cost_function}

\begin{table}[H]
\small\sf\centering
\begin{adjustbox}{max width=\columnwidth}
\begin{tabular}{@{}lllllll@{}}
\hline
$q_x$ & $q_v$ & $q_{\theta}$ & $q_{\dot{\theta}}$ & $R$ & $R_d$ & $T~[\si{\second}]$ \\
\hline
$0$ & $500$ & $328.28$ & $0$ & $0$ & $50$ & $0.41$ \\
\hline
\end{tabular}
\end{adjustbox}
\end{table}

\subsection{SMB Actuator Model Evaluation}
\label{tab:smb_act_model_eval}

The measurement rate is 100~\si{\hertz}, and the actuator model predicts the torque at the same rate. The observation histories consist of 30 datapoints and cover the past 0.3~\si{\second}. The training and validation datasets are recorded by driving the \ac{SMB} forth and back with varying speeds on flat ground. Before training the model, the datasets were standardized by subtracting the mean and dividing by the standard deviation. In the following table, we report the Mean average error (Mae), Standard deviation (Std), 95th percentile (p95), and Delay Estimation (Lag) computed using normalized cross-correlation, between the measured base velocity in simulation and reality. The Lag between the command and the measured base velocity corresponds to 0.26~\si{\second}.

\begin{table}[H]
\small\sf\centering
\begin{adjustbox}{max width=\columnwidth}
\begin{tabular}{@{}llllll@{}}
\hline
\textbf{Dataset} & \textbf{Duration $[\si{\second}]$} & \textbf{Mae $[\si{\m\per\second}]$} & \textbf{Std $[\si{\m\per\second}]$} & \textbf{p95 $[\si{\m\per\second}]$} & \textbf{Lag $[\si{\second}]$} \\
\hline
Training & 320 & 0.040 & 0.053 & 0.115 & 0.020 \\
Validation & 290 & 0.040 & 0.054 & 0.117 & 0.030 \\
\hline
\end{tabular}
\end{adjustbox}
\end{table}

\subsection{SMB Policy Observations}
\label{tab:smb_observed_variables}

The controller is deployed at a rate of 50~\si{\hertz}. The noise is sampled from a Gaussian distribution with mean $\mu = 0$ and standard deviation $\sigma$. The \ac{IMU} is a MicroStrain MV5-AR~\cite{smb_imu} and the wheel encoders are AMT10 by Same Sky~\cite{smb_encoder}.

\begin{table}[H]
\small\sf\centering
\begin{adjustbox}{max width=\columnwidth}
\begin{tabular}{@{}llll@{}}
\hline
\textbf{Observation} & \textbf{History} & \textbf{Stride} & \textbf{Noise $\sigma$} \\
\hline
\ac{IMU} acceleration $(x,y,z)$ $[\si{\m\per\second\squared}]$ & 15 & 1 & 0.01 \\
\ac{IMU} angular velocity $(\omega_x, \omega_y, \omega_z)$ $[\si{\radian\per\second}]$ & 15 & 1 & 0.02 \\
Wheel speeds $[\si{\radian\per\second}]$ & 15 & 1 & 0.1 \\
Pitch $[\si{\radian}]$ & 15 & 1 & 0.0 \\
Prev. action $[\si{\percent}]$ & 15 & 1 & 0.0 \\
Target lin. vel. $[\si{\m\per\second}]$ & 15 & 1 & 0.0 \\
Target ang. vel. $[\si{\radian\per\second}]$ & 15 & 1 & 0.0 \\
\hline
\end{tabular}
\end{adjustbox}
\end{table}

\subsection{SMB Rigid-Body Model Randomization}
\label{tab:smb_param_rand}

The rigid-body randomization parameters are sampled from a uniform distribution within $\pm$ the values listed in this table.
\begin{table}[H]
\small\sf\centering
 \begin{adjustbox}{max width=\columnwidth}
 \begin{tabular}{@{}l l@{}}
 \hline
 \textbf{Parameter} & \textbf{Value} \\
\hline
 Mass $[\si{\percent}]$ & 5 \\
 Inertia $\Theta ~[\si{\percent}]$ &  20 \\
 \acs{CoM} $x,y,z~[\si{\m}]$  & 0.05 \\
\hline
 \end{tabular}
\end{adjustbox}
\end{table}

\subsection{SMB Actuator Perturbation}
\label{tab:smb_act_noise}

An external disturbance is implemented as actuator noise. The magnitude is sampled from a uniform distribution within $\pm$ the maximum torque and resampled at the given resampling time.

\begin{table}[H]
\small\sf\centering
  \begin{adjustbox}{max width=\columnwidth}
 \begin{tabular}{@{}l l@{}}
 \hline
 \textbf{Parameter}  & \textbf{Value} \\
 \hline
 Maximum torque $[\si{\newton\meter}]$& 5.0 \\
 Resampling time $[\si{\second}]$& 1.0 \\
 \hline
 \end{tabular}
 \end{adjustbox}
\end{table}

\subsection{SMB Reward Function Coefficients}
\label{tab:smb_reward}

 $c_3$ and $c_4$ are increased exponentially with the curriculum multiplier $\mu$ as follows: $c_{3/4}\leftarrow c_{3/4}(1+\mu)$.

\begin{table}[H]
\small\sf\centering
\begin{adjustbox}{max width=\columnwidth}
\begin{tabular}{@{}r l@{}}
    \hline
    \textbf{Coefficient} & \textbf{Value} \\
    \hline
    $c_1$ & 0.01 \\
    $c_2$ & 0.001 \\
    $c_3$ & 0.001 \\
    $c_4$ & 0.001 \\
    $c_{3,\textrm{max}}$ & 0.006 \\
    $c_{4,\textrm{max}}$ & 0.006 \\
    $\mu$ & 0.006 \\
    \hline
\end{tabular}
\end{adjustbox}
\end{table}

\subsection{SMB Policy Training Parameters}
\label{tab:smb_training}

Parameters for training the policy.

\begin{table}[H]
\small\sf\centering
 \begin{adjustbox}{max width=\columnwidth}
 \begin{tabular}{@{}l l@{}}
 \hline
 \textbf{Parameter} & \textbf{Value} \\
 \hline
 Policy net. & 256x128  \\
 Critic net. & 256x128  \\
 Nonlinearity & tanh  \\
 \Ac{PPO} iterations & 1000 \\
 Batch size & 100k \\
 Episode length [\si{\second}] & 10 \\
 Control rate [\si{\hertz}] & 50 \\
  Discount factor & 0.998 \\
 Entropy coeff. & 0.0 \\
 Learning rate & 1e-3 \\
 Epochs & 5 \\
 Mini-batches & 2 \\
\hline
\end{tabular}
\end{adjustbox}
\end{table}

\end{document}